\documentclass[conference]{IEEEtran}
\usepackage{times}

\usepackage[numbers]{natbib}
\usepackage{multicol}
\usepackage[bookmarks=true]{hyperref}

\usepackage{siunitx}

\usepackage{amsmath,amssymb,amsfonts}
\usepackage{algorithmic}
\usepackage{graphicx}
\usepackage{textcomp}
\usepackage{setspace}
\usepackage{booktabs}
\usepackage{booktabs}
\usepackage{multirow}
\usepackage[table]{xcolor} 

\newcolumntype{g}{>{\columncolor{gray!30}}r}

\usepackage{epsfig}
\usepackage{times} 
\usepackage{svg}
\usepackage[linesnumbered,ruled,vlined]{algorithm2e} 

\usepackage[nolist, nohyperlinks]{acronym}
\acrodef{GNC}[GNC]{Graduated-Nonconvexity}
\acrodef{WLS}[WLS]{Weighted Least Squares}
\acrodef{TLS}[TLS]{Truncated Least Squares}
\acrodef{SVD}[SVD]{Singular Value Decomposition}
\acrodef{VO}[VO]{Visual Odometry}
\acrodef{VIO}[VIO]{Visual Inertial Odometry}
\acrodef{ESDF}[ESDF]{Euclidean Signed Distance Field}
\acrodef{CESDF}[C-ESDF]{Certified \ac{ESDF}}
\acrodef{TSDF}[TSDF]{Truncated Signed Distance Field}
\acrodef{RGBD}[RGBD]{RGB-Depth}
\acrodef{IMU}[IMU]{Inertial Measurement Unit}
\acrodef{KF}[KF]{Kalman Filter}
\acrodef{EKF}[EKF]{Extended Kalman Filter}
\acrodef{VO}[VO]{Visual Odometry}
\acrodef{CVO}[C-VO]{Certified Visual Odometry}
\acrodef{ISS}[ISS]{Input-to-State}
\acrodef{SLAM}[SLAM]{Simultaneous Localization and Mapping}
\acrodef{SFC}[SFC]{Safe Flight Corridor}
\acrodef{FOV}[FoV]{Field of View}
\acrodef{BCH}[BCH]{Baker-Campbell-Hausdorff}
\acrodef{FPV}[FPV]{First Person View}

\usepackage{bm}
\usepackage{breqn}
\usepackage{amsmath}
\usepackage{amssymb}
\usepackage{tabularx}

\usepackage{amsthm}

\newcommand{\naturals}{\mathbb{N}}

\newcommand{\reals}{\mathbb{R}}
\newcommand{\R}{\reals}
\newcommand{\Rnonneg}{\reals_{\geq 0}}
\newcommand{\Rplus}{\reals_{>0}}

\renewcommand{\S}{\mathbb{S}}
\newcommand{\SO}{\mathbb{SO}}
\newcommand{\SE}{\mathbb{SE}}
\newcommand{\so}{\mathfrak{so}}
\newcommand{\se}{\mathfrak{se}}

\newcommand{\pd}{\S_{+}}

\newcommand{\Ecal}{\mathcal{E}}
\newcommand{\Fcal}{\mathcal{F}}

\newcommand{\Hcal}{\mathcal{H}}

\newcommand{\Mcal}{\mathcal{M}}
\newcommand{\Ncal}{\mathcal{N}}
\newcommand{\Ocal}{\mathcal{O}}
\newcommand{\Pcal}{\mathcal{P}}

\newcommand{\Rcal}{\mathcal{R}}
\newcommand{\Scal}{\mathcal{S}}

\newcommand{\Ucal}{\mathcal{U}}
\newcommand{\Vcal}{\mathcal{V}}
\newcommand{\Wcal}{\mathcal{W}}

\newcommand{\eqn}[1]{\begin{align} #1 \end{align}}
\newcommand{\eqnN}[1]{\begin{align*} #1 \end{align*}}
\newcommand{\neqn}[1]{\begin{align*} #1 \end{align*}}
\newcommand{\seqn}[2][]{
\begin{subequations}
    \label{#1}
\begin{align} #2 \end{align}
\end{subequations}
}

\newcommand{\bmat}[1]{\begin{bmatrix}#1\end{bmatrix}}

\newcommand{\edit}[1]{#1}
\newcommand{\norm}[1]{\left\Vert #1 \right \Vert}

\newcommand{\inframe}[1]{|^{#1}}

\newcommand{\T}[2]{T_{#1}^{#2}}
\newcommand{\estT}[2]{\widehat{T}_{#1}^{#2}}
\newcommand{\Exp}{\operatorname{Exp}}
\newcommand{\Log}{\operatorname{Log}}
\newcommand{\Ad}[1]{\operatorname{Ad}_{#1}}

\theoremstyle{plain}
\newtheorem{theorem}{Theorem}

\newtheorem{lemma}{Lemma}
\newtheorem{problem}{Problem}
\newtheorem{definition}{Definition}

\newtheorem{assumption}{Assumption}
\newtheorem{remark}{Remark}

\theoremstyle{remark}

\usepackage{cleveref}

\crefformat{problem}{Problem~#2#1#3}
\crefformat{assumption}{Assumption~#2#1#3}

\usepackage[absolute,overlay]{textpos}

\newcommand*\pct{\scalebox{.85}{\%}}

\begin{document}

\title{Certifiably-Correct Mapping for Safe Navigation \\ Despite Odometry Drift}




%

\author{
Devansh R. Agrawal, Taekyung Kim, Rajiv Govindjee, Trushant Adeshara,\\
 Jiangbo Yu, Anurekha Ravikumar, and Dimitra Panagou\\
University of Michigan, Ann Arbor\\
Correspondence: \texttt{devansh@umich.edu}
}


\maketitle

\begin{abstract}
Accurate perception, state estimation and mapping are essential for safe robotic navigation as planners and controllers rely on these components for safety-critical decisions. However, existing mapping approaches often assume perfect pose estimates, an unrealistic assumption that can lead to incorrect obstacle maps and therefore collisions. This paper introduces a framework for certifiably-correct mapping that ensures that the obstacle map correctly classifies obstacle-free regions despite the odometry drift in vision-based localization systems (\acs{VIO}/\acs{SLAM}). By deflating the safe region based on the incremental odometry error at each timestep, we ensure that the map remains accurate and reliable locally around the robot, even as the overall odometry error with respect to the inertial frame grows unbounded.

Our contributions include two approaches to modify popular obstacle mapping paradigms, (I)~Safe Flight Corridors, and (II)~Signed Distance Fields. We formally prove the correctness of both methods, and describe how they integrate with existing planning and control modules. Simulations using the Replica dataset highlight the efficacy of our methods compared to state-of-the-art techniques. Real-world experiments with a robotic rover show that, while baseline methods result in collisions with previously mapped obstacles, the proposed framework enables the rover to safely stop before potential collisions.

\end{abstract}

\IEEEpeerreviewmaketitle

Code\footnote{Code: \url{https://github.com/dasc-lab/certifiably-correct-mapping}} and Video\footnote{Video: \url{https://youtu.be/qMlDK7Iou48}}

\section{Introduction}
\label{sec:introduction}

\begin{textblock*}{15cm}(2cm,1cm) 
	Authors Copy. This paper has been accepted for publication in RSS 2025. 
\end{textblock*}

\acresetall

Accurate state estimation and mapping are essential for safe robotic navigation, as planners and controllers rely on perception outputs to ensure the safety of planned trajectories or control actions. Various methods have been developed to certify that controllers meet predefined safety specifications~\cite{ames2016control, garg2023advances}, and when real-time obstacle detection is necessary, it is often intuitive to handle safety constraints in the planner~\cite{lopez2017aggressive, tordesillas2019faster, agrawal2024gatekeeper}. These methods typically assume perfect perception, a simplification that can lead to safety violations.

A perception module provides a pose estimates and constructs maps of the obstacle geometry, and can take a variety of formats, such as \acp{ESDF}~\cite{oleynikova2017voxblox, nvblox}, polytopic \acp{SFC}~\cite{liu2017planning}, occupancy log-odds~\cite{hornung2013octomap}, or NERFs~\cite{rosinol2023nerf}. Although recent advances have achieved significant accuracy improvements~\cite{scaramuzza2011visual, yu2021vins, tian2022kimera, chen2023direct, merat2025drift}, formal error analysis is often lacking. Without quantified error bounds, guaranteeing the safety of a closed-loop robotic system remains a challenge.

\begin{figure}
    \centering
    \includegraphics[width=0.97\linewidth]{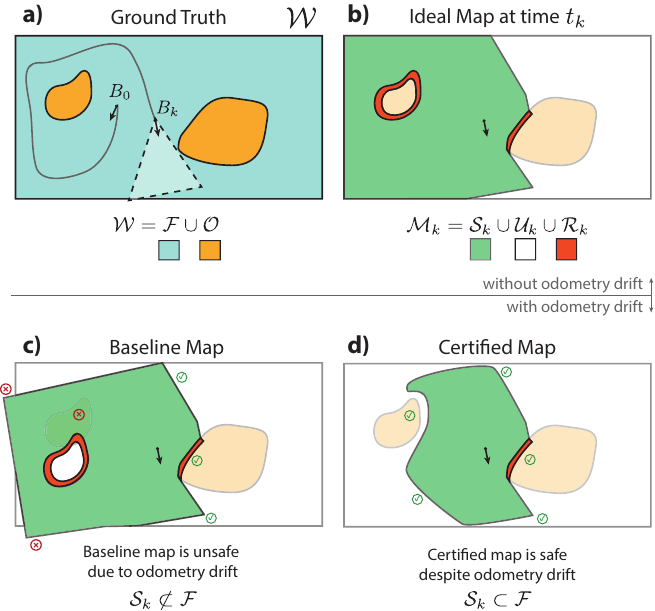}
    \caption{Overview of notation and objectives. (a) depicts the operating environment, where the world $\Wcal$ is the union of the free space $\Fcal$ and the obstacles $\Ocal$. The robot does not know $\Fcal$ or $\Ocal$. It starts at $B_0$, and follows the gray trajectory to $B_k$ building the map as it goes. (b) depicts the ideal mapping output, where at the $k$-th timestep, the map $\Mcal_k$ is composed of the known safe region $\Scal_k$, the unknown space $\Ucal_k$ and the known obstacle set $\Rcal_k$. (c) depicts the map produced by current state-of-the-art methods, where due to odometry drift the map is erroneous: notice that the safe region (according to the constructed map) is not a subset of the free space, $\Scal_k \not\subset \Fcal$. (d) depicts the desired behavior of the certified maps, where although the safe region is smaller, it is certifiably-correct: we can prove that $\Scal_k \subset \Fcal$. }
    \label{fig:notation}
\end{figure}

This paper introduces a framework for ``certifiably correct mapping" ensuring that obstacle-free regions of a map remain correct despite odometry drift. The challenge is illustrated in~\Cref{fig:notation}.  Consider an environment $\Wcal = \Fcal \cup \Ocal$, representing free and obstacle spaces, respectively~(\Cref{fig:notation}a). As a robot navigates,  at the $k$-th time step it has created a map $\Mcal_k$, comprising the supposedly safe space $\Scal_k$, the unknown space $\Ucal_k$ and the recognized obstacles $\Rcal_k$~(\Cref{fig:notation}b). However, due to odometry drift, maps can misclassify obstacles as free space, leading to potential safety violations as indicated in~\Cref{fig:notation}c. We address this by deflating safe regions in order to ensure $\Scal_k \subset \Fcal$ at all times~(\Cref{fig:notation}d).

Our main contributions are as follows: 
\begin{itemize}
    \item The theoretical framework to construct and deflate the free space in obstacle maps to ensure their correctness despite odometry drift.  Assuming the odometry algorithm reports the pose and the covariance of the incremental transform, we propose  deflating the supposedly safe region ($\Scal_{k+1}$ is deflated relative to $\Scal_k$) to ensure that it remains a subset of the free region $\Fcal$.
    \item We prove the correctness and applicability of this framework on two popular and state-of-the-art mapping frameworks: the polytopic~\acp{SFC} of~\cite{liu2017planning} and the \acp{ESDF} of~\cite{nvblox}.
    \item Beyond providing the theoretical analysis and proofs of correctness, we validate and compare our approach with state-of-the-art baseline methods through extensive simulations on the Replica dataset~\cite{replica19arxiv}. 
    \item Finally, we demonstrate the approach in a real-world experiment on a robotic rover. A human teleoperates the rover using only the \ac{FPV} feed and the obstacle map constructed and streamed to the operator in real-time. The rover uses an onboard safety filter to prevent collisions. Unlike baseline methods which result in collisions,  our approach prevents crashes by deflating the safe regions appropriately. 
\end{itemize}

It is critical that we deflate $\Scal_k$ rather than inflate known obstacles $\Rcal_k$. If the obstacles are inflated based on the accumulated odometry error, these obstacles can only grow in size, and might eventually occupy the entire domain $\Wcal$. Instead, by deflating a safe region $\Scal_k$, the region that is certifiably safe shrinks, eventually becomes an empty set, and is removed from memory (i.e., the region becomes part of $\Ucal_k$). When the region is observed by a sensor again, it can again be added to $\Scal_k$ again. Computationally, this reduces memory requirements, and mathematically this allows us to treat deflated obstacles as unknown regions and plan paths accordingly. \edit{The certified maps can be used together with the uncertified maps for practical applications: the uncertified maps can be used to plan trajectories for example for exploration or for navigating towards a goal location, while the certified map can be used for local obstacle avoidance.}

Our paper is organized as follows. After a brief literature review in~\Cref{section:lit_review}, in~\Cref{sec:notation} we provide a mathematical background and setup the problem formally. In~\Cref{section:certified_sfc} and~\ref{section:certified_sdf} we introduce the deflation mechanism for both map representations. In~\Cref{section:safe_navigation} we propose methods to use the certified maps to acheive safe navigation. Finally in~\Cref{section:simulations} and~\Cref{section:experiments} we present the simulation and experimental results. 

\edit{
\section{Literature Review}
\label{section:lit_review}

Perception methods have seen significant advancements over the past few decades, driven by improvements in algorithms, sensors, and computational capabilities~\cite{cadena2016past, macario2022comprehensive}. The primary goals of these advancements have been to enhance localization and mapping accuracy, improve robustness under diverse environmental conditions, and develop algorithms with lower computational costs. For instance, modern \ac{SLAM} systems now report translation error rates below 1\%~\cite{vslam, campos2021orb}, enabling more reliable navigation in real-world scenarios.

With these improvements, robots have been deployed in increasingly complex environments, relying heavily on \ac{VIO}/\ac{SLAM} pose estimates and obstacle maps to navigate safely. As exemplified by the DARPA SubT Challenge, teams have developed perception systems capable of navigating subterranean environments~\cite{ebadi2023present, chung2023into, tranzatto2022cerberus}. In these systems, raw measurements are typically processed by a frontend into a more compact representation, while a backend uses nonlinear optimization methods to compute the robot's trajectory and map estimate~\cite{ebadi2023present}. Most of these optimization methods are based on factor graphs, which, although effective, become computationally expensive as the map size increases.

A common approach to manage this computational complexity is to use local submaps, connected through a graph of traversable regions or submap connections~\cite{ebadi2023present}. These methods reduce odometry drift by optimizing each submap within its own coordinate frame. When a robot revisits a previously mapped region, the submap can be reused, provided that the robot is correctly localized within it. However, even within a submap, odometry drift can still lead to localization errors. Therefore, ensuring safety requires addressing the potential errors within these submaps. The approach proposed in this paper aims to ensure correctness at the submap level, i.e., in the presence of incremental localization errors.

Recent work has explored techniques for ensuring the correctness of perception systems. For example, \cite{rosen2019se}~achieve global optimization in pose graph optimization problems through a convex reformulation, while~\cite{marchi2022lidar} provide error-bounded localization within 2D convex environments. Additionally,~\cite{yang2020teaser, agrawal2024online} propose certifiably correct point-cloud registration and visual odometry methods. Similarly,~\cite{zhang2015ins} showed that bounded attitude errors lead to bounded position errors. In contrast to~\cite{agrawal2024online}, this paper assumes that the incremental pose estimate is bounded in a Lie-algebraic sense, which allows our methods to be applied to a broader range of odometry algorithms, extending the applicability beyond the methods considered in~\cite{agrawal2024online}. In cases where certification of correctness is not feasible, estimating or quantifying the error can still provide valuable insights, for example using the methods in~\cite{maken2021stein, laconte2023toward} which estimate the error in point-cloud matching.

Other approaches have been proposed to address mapping consistency in the presence of odometry drift. \cite{millane2018c}~utilize overlapping \ac{TSDF} voxels, which are only fused once the consistency of certain regions has been verified. These ideas share similarities with the work of~\cite{howard2006multirobot, cieslewski2019exploration}, which also emphasize the importance of ensuring consistency before merging obstacle estimates from different times. These methods propose constructing a manifold map, only merging them when correctness can be guaranteed. In contrast, the method proposed in this paper introduces a different strategy: regions where correctness cannot be assured are "forgotten," ensuring that only reliable, consistent parts of the map are used for navigation and decision-making.

}

\section{Preliminaries and Problem Statement}
\subsection*{Notation}
\label{sec:notation}

$\naturals= \{0, 1, 2, ...\}$ is the set of natural numbers. 
$\R, \Rnonneg, \Rplus$ denote reals, non-negative reals, and positive reals.
$I_n \in \R^{n \times n}$ is the $n \times n$ identity matrix. The subscript is dropped when clear from context.
$\SO(n)$ is the $n$-d special orthogonal group. 
$\SE(n)$ is the $n$-d special Euclidean group. 
$\pd^n$ is the set of symmetric positive-definite matrices in $\R^{n \times n}$.  
The matrix square root of positive definite matrix $A \in \pd^{n}$ is the matrix $A^{1/2} \in \R^{n \times n}$ such that $A^{1/2} A^{1/2} = A$. 
For $v\in \R^n$, $\norm{v}$ denotes the 2-norm, $\norm{v}_p$, $(p\in [1, \infty])$ denotes the $p$-norm, and $\norm{v}_P = \sqrt{ v^T P v}$ for $P \in \pd^n$. 
All eigenvectors are assumed to be unit-norm. $\lambda(A)$ is the set of eigenvalues of $A \in \R^{n \times n}$, and $\lambda_{\max}(A)$ is the largest eigenvalue of $A \in \pd^n$. 
$[p]_\times \in \R^{3 \times 3}$ is the skew-symmetric matrix such that $a \times b = [a]_\times b$ for any $a, b \in \R^3$. 


\subsection*{Matrix Lie Groups}

Here we present a brief review of Matrix Lie Groups in the context of this paper, with additional equations and details in~\Cref{appendix:lie_groups}. We refer the reader to the excellent references~\cite{sola2018micro, mangelson2020characterizing, barfoot2024state} for a more complete description. 

The Lie group $\SO(3)$ defines 3D rotations, and the group $\SE(3)$ defines 3D rigid transformations. Both $\SO(3)$ and $\SE(3)$ are Matrix Lie groups, i.e., group elements are matrices, and composition operator is the standard matrix multiplication operator.  In $\SE(3)$ the group action $\cdot : \SE(3) \times \R^3 \to \R^3$ transforms a point $p$ from its representation in frame $A$ to that in frame $B$. Given $\T{A}{B} = \bmat{R & t \\ 0 & 1} \in \SE(3)$, 
\eqn{
p\inframe{B} = \T{A}{B} \cdot p\inframe{A} = R p \inframe{A} {} + {}  t.
}

The Lie algebra of a group is a vector space of all possible directions a group element can be perturbed locally. The Lie algebras of $\SO(3)$ and $\SE(3)$ are $\so(3)$ and $\se(3)$ respectively. These vector spaces are isomorphic to $\R^3$ and $\R^6$ respectively. The $\wedge$ operator converts the Euclidean vector to an element of the Lie Algebra, and $\vee$ does the inverse. 

Consider a Lie group $\mathbb{G}$ with an associated Lie algebra $\mathfrak{g}$ that is isomorphic to the Euclidean vector space $\R^n$. Given an element $x \in \mathfrak{g}$, we can convert it to the corresponding group element using the exponential map, $\exp: \mathfrak{g} \to \mathbb{G}$.
For convenience, we also define the $\Exp$ map, which maps from the Euclidean vector space to the group directly, $\Exp: \R^n \to \mathbb{G}$,
$
\Exp(\xi) = \exp(\xi^\wedge).
$
For certain groups including $\SE(3)$, these operations have analytic expressions~\cite[Appendix]{sola2018micro}.

\subsection*{Uncertain Poses and Transforms}

An uncertain pose or transform $\T{A}{B} \in \SE(3)$ is denoted 
\eqnN{
\T{A}{B} \sim \Ncal(\estT{A}{B}, \Sigma_T),
} 
where $\estT{A}{B} \in \SE(3)$ is the mean estimate, and $\Sigma_T \in \pd^6$ is a covariance matrix. This indicates $\T{A}{B}$ is the transform
\eqn{
\T{A}{B} = \estT{A}{B} \Exp{\tau},
}
where $\tau \in \R^6$ is a random sample drawn from $\tau \sim \Ncal(0, \Sigma_T)$.

Recall the group action $p\inframe{B} = \T{A}{B} \cdot p\inframe{A}$. If the transform $\T{A}{B}$ is uncertain, $p\inframe{B}$ follows a distribution and, to first order, is a normal distribution~\cite{sola2018micro, barfoot2024state}:
\eqn{
p\inframe{B} = \left(\T{A}{B} \cdot p\inframe{A}\right) \sim \Ncal( \hat p\inframe{B}, \Sigma_p)
}
where the mean and covariance are 
\eqnN{
\hat p \inframe{B} &= \estT{A}{B} \cdot p\inframe{A} \in \R^3, \quad \Sigma_p = J \Sigma_T J^T \in \pd^{3}
}
with $J  = \bmat{ R & -R [p\inframe{A}]_\times} \in \R^{3 \times 6}$. 

For the remainder of the paper, we truncate the distribution making the following assumption:

\begin{assumption}
        \label{assumption:p}
        Let $\T{A}{B} \sim \Ncal(\estT{A}{B}, \Sigma)$, where $\estT{A}{B} = \bmat{R & t \\ 0 & 1}$.
        Then for any $p\inframe{A} \in \R^3$, the point $p\inframe{B} \in \R^3$ satisfies
        \eqn{
            p\inframe{B} = \T{A}{B} \cdot p\inframe{A} \in \Ecal
        }
        where $\Ecal \subset \R^3$ is the ellipsoid
        \eqn{
            \Ecal &= \left\{ p \in \R^3: \norm{ \Sigma_p^{-1/2} \left(p - \estT{A}{B} \cdot p\inframe{A} \right)} \leq 1 \right\}, \label{eqn:ellipsoid}
        }
        \eqnN{
        \Sigma_p = \kappa J \Sigma J^T \in \pd^3, \quad 
        J = \bmat{ R & -R [p \inframe{A}]_\times}  \in \R^{3 \times 6}.
        } 
        for some $\kappa>0$.\footnote{$\kappa$ chooses the probability the bound contains the point. For a $d$-dimensional normal distribution, $x \sim \Ncal(\mu, \Sigma)$, the probability that $\norm{ (\kappa \Sigma)^{-1/2}(x - \mu)} \leq 1$ is $p \in [0, 1]$ such that $\kappa = \chi^2_d(p)$, where $\chi^2_d$ is the quantile function of the chi-squared distribution with $d$ degrees of freedom. For 3D points,  $\kappa=2$ corresponds to $p=97\%$.}
\end{assumption}

In other words, the assumption is that when a point $p$ is transformed from its representation in frame $A$ to that in frame $B$, the point $p\inframe{B}$ is contained within an ellipsoid $\Ecal$ centered on the estimated point $\estT{A}{B} \cdot p\inframe{A}$, as defined in~\eqref{eqn:ellipsoid}. The size and principal axes of the ellipsoid are defined by the estimated transform $\estT{A}{B}$ and the covariance matrix $\Sigma$. This allows us to bound the error of mapping points between frames, and the bound can be made tighter if $\kappa$ is increased, or if higher order approximations are used, as in~\cite{barfoot2024state}. \edit{The higher order approximations yield tighter covariance ellipsoids, at the expense of increased computation.} Since we focus on rototranslations between successive body frames, the transforms $\T{A}{B}$ should be close to identity where first order approximations work well.

\subsection*{Reference Frames}

This paper uses the inertial frame $I$, a mapping frame $M$, and the body-fixed frame at the $k$-th timestep, $B_k$. Usually, $M$ and $I$ are equivalent, and $M$ is defined such that at $M = B_0$. However, since we are considering odometry drift, $M$ can drift relative to $I$. We assume that $I$ is the true inertial frame (in which the obstacles are static), and $M$ is the reference frame used to construct the state estimate and the map. 

\subsection*{Problem Statement}
\label{section:problem_statement}

Let $\Ocal$ represents the obstacle geometry in a static environment $\Wcal \subset \R^3$.  Both $\Ocal$ and  $\Fcal=\Wcal \backslash \Ocal$ are assumed initially unknown. \edit{We assume $\Fcal$ does not contain any isolated points, and that $\Ocal$ is closed.} As with points, a set can be represented in a frame, i.e., we say that $\Ocal\inframe{B_k} \subset \R^3$ is the set of all obstacle points represented in frame $B_k$.

To avoid obstacles, we must build a map of the environment. At the $k$-th timestep the map is $\Mcal_k$, consisting of the (claimed) free-space $\Scal_k$, the unknown space $\Ucal_k$, and the (claimed) obstacle space $\Rcal_k$.  A map is correct if the claimed free space is a subset of the true free space.\footnote{Since $\Ocal$ is closed, $\Fcal$ is open. The (claimed) safe region $\Scal$ can be either an open or closed subset of $\Fcal$. Below, $\Scal$ will be a closed set.} More formally, 
\begin{definition}
\label{defintion:correct_map}
        A map $\Mcal = \Scal \cup \Ucal \cup \Rcal$ is the union of the (claimed) safe region $\Scal$, the unknown region $\Ucal$, and the (claimed) obstacle region $\Rcal$. At the $k$-th timestep, the map $\Mcal_k$ is \emph{correct} if for all $p\inframe{B_k} \in \R^3$, 
        \eqn{
                p\inframe{B_k} \in \Scal_k\inframe{B_k} \implies p\inframe{B_k} \in \Fcal\inframe{B_k}.
        }
\end{definition}
In words, $\Mcal_k$ is \emph{correct} if $\Scal_k$ is a subset of the free space $\Fcal$ \emph{when represented in the $k$-th body-fixed frame}.

The definition above is intentionally explicit about which reference frame various points and sets are represented in since this is the source of the main problem tackled in this paper. 
Due to the odometry drift, there are two types of error common in state-of-the-art mapping algorithms:

\emph{(A) Errors in constructing the map:}
In current state-of-the-art implementations, the map is often represented computationally in the mapping frame $M$. Suppose at some time $t_k$ the robot detects an obstacle (relative to its body-fixed camera) at a position $p\inframe{B_k}$. It will update the map to remove this point from the claimed free space:
\eqn{
\label{eqn:source_of_error_A}
\Scal_{k+1}\inframe{M} \subset \Scal_{k}\inframe{M} \backslash \{ \estT{B_k}{M} \cdot p\inframe{B_k} \}.
}
However, notice that since the estimated transform $\estT{B_k}{M}$ is used instead of the true transform $\T{B_k}{M}$, the location marked as an obstacle can be wrong. This problem is exacerbated since usually the line connecting the camera origin and the point $p\inframe{B_k}$ is marked free, and therefore the wrong locations are marked as part of $\Scal_{k+1}$.

\emph{(B) Errors in querying the map: }
Now suppose the robot wants to navigate the environment. It must therefore (at time $t_k$) check whether a point $p\inframe{B_k}$ relative to the body-fixed frame is free. To the best of our knowledge, all implementations will then check whether the corresponding estimated point in the map, $\hat p \inframe{M}$, is a free point, that is, they check whether
\eqn{
\label{eqn:source_of_error_B}
\hat p\inframe{M} = \estT{B_k}{M} \cdot p\inframe{B_k} \  \in \Scal_k\inframe{M}.
}
However notice again, since the estimated transform is used, this can lead to inconsistencies. In particular, owing to the odometry drift, the inconsistency will be worse when the obstacle point was inserted into the map many frames ago.\footnote{It will also becomes clear that time is not the only factor - points inserted/queried further from the robot will also be more inaccurate due to the larger moment arm that amplifies rotation errors. This is also why common heuristic algorithms of time- or distance-based forgetting cannot guarantee the correctness of the map. The methods proposed in this paper will directly address such issues. }

We overcome both such issues, \emph{by ensuring the map is always correct in the body-fixed frame.} An equivalent perspective is that despite using the estimated transform $\estT{B_k}{M}$ the map will be constructed and queried correctly.

The problem statement therefore is as follows:
\begin{problem}
    Consider a robotic system equipped with an RGBD camera operating in a static environment with obstacles $\Ocal \subset \R^3$.
    Suppose an odometry module provides at each frame $k$ the estimated odometry $\estT{B_k}{B_0} \in \SE(3)$, the relative odometry $\estT{B_{k+1}}{B_{k}} \in \SE(3)$ and a covariance of the relative odometry $\Sigma_{B_{k+1}}^{B_k} \in \pd^6$. 
    Suppose a mapping module can construct the best estimate map of the free space in the environment. 
    Design a framework to correct the best-estimate map such that at each timestep, the map $\Mcal_k$ is correct according to \Cref{defintion:correct_map} \emph{despite the odometry drift}. 
\end{problem}

We also assume that if an obstacle point is within the camera's \ac{FOV}, it will be detected as an obstacle. This is a common implicit assumption in the mapping literature. Infrared depth cameras often fail to detect transparent obstacles (e.g., windows and glass doors) or obstacles with minimal texture (where the stereo block-matching algorithm fails). Such issues are beyond the scope of this paper.

\begin{figure*}[t]
    \centering
    \includegraphics[width=0.9\linewidth]{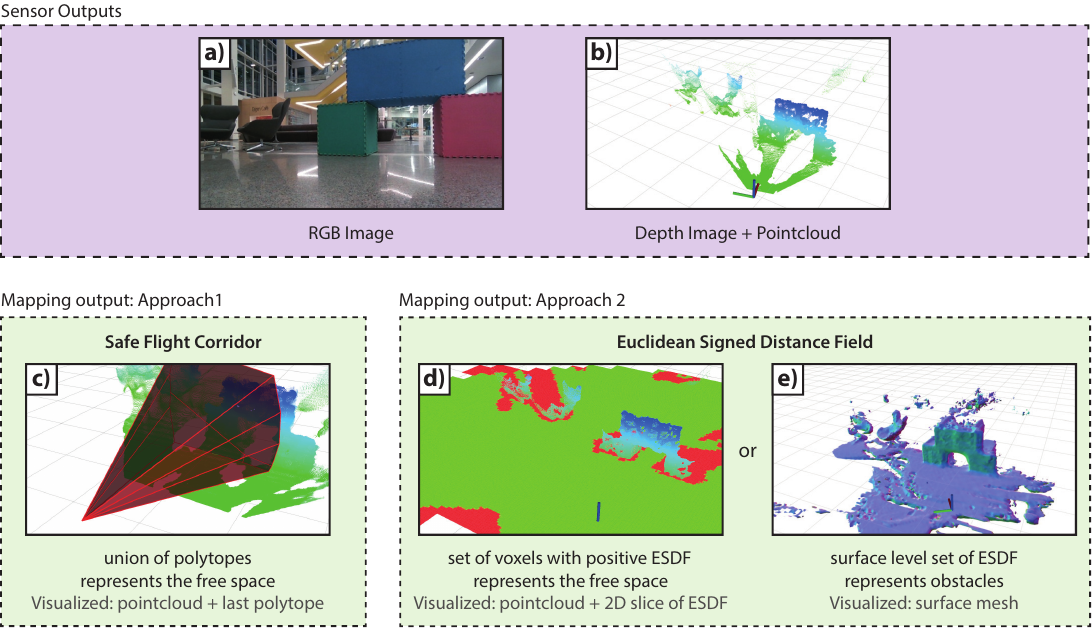}
    \caption{Two approaches to constructing an obstacle map. (Top row) An \ac{RGBD} camera provides (a)~the first person RGB image, and (b)~the depth image/pointcloud constructed from stereo images. (Bottom row) The \ac{SFC} approach represents the free space as a union of polytopes, one of which is depicted in~(c). The \ac{ESDF} approach represents the world using voxels, where each voxel stores the signed distance to the nearest obstacle. From this, both the (d)~\ac{ESDF} at specific voxels or (e)~obstacle surface locations can be extracted and used for safe navigation. To aid the reader, in (c) and (d) the raw pointcloud is also visualized, and in (d)~the colorscheme is such that voxels are marked green if $d > 0$, and red otherwise. This makes the map look binary, although it contains continuous values. Furthermore, note both methods operate in 3D - the 2D slice is used for visualization. }
    \label{fig:mapping_outputs}
\end{figure*}

In the next two sections, we demonstrate how to construct correct maps by modifying existing baseline mapping algorithms. In particular we extend (A)~a mapping algorithm~\cite{liu2017planning} which uses polytopes to represent the map of free space, and (B)~the mapping algorithm~\cite{nvblox} which uses signed distance fields to represent the free space. See \Cref{fig:mapping_outputs}.

\section{Approach 1: Certified Safe Flight Corridors}
\label{section:certified_sfc}

\subsection{Background}

In the first approach, the obstacle-free region $\Scal_k$ at frame $k$ is the union of $n$ polytopes,\footnote{$n$ can be different at each $k$. }
\eqn{
\label{eqn:Mcal_k}
        \Scal_k\inframe{B_k} = \bigcup_{l=1}^{n} \Pcal_k^l 
}
where each polytope is a compact set of the form
\eqn{
        \Pcal_k^l = \{ p \in \R^3 : A_k^l p \leq b_k^l \}.
}
This is often called the H-representation, since the polytope is defined by a set of half-space constraints~\cite{legat2023polyhedral}. An example of a polytope extracted from a depth image is shown in~\Cref{fig:mapping_outputs}c. 

As the robot transitions from frame $B_k$ to frame $B_{k+1}$, we can map each polytope from the previous frame to the new frame, and maintain the polytopes in the robot's body frame. 

\emph{In the absence of odometry drift}, one can directly compute the new polytopes:
\seqn[eqn:basic_polytope]{
        \Pcal_{k+1}^l &= \{ p \in \R^3 : A_{k+1}^l p \leq b_{k+1}^l \},\\
        A_{k+1}^l &= A_k^l R^T,\\
        b_{k+1}^l &= b_k^l + A_k^l R^T t,
}
using the estimated transforms
\eqnN{
        \estT{B_k}{B_{k+1}} = \bmat{R & t \\ 0& 1 }.
}

In the presence of odometry drift, however, the estimated transform $\estT{B_{k}}{B_{k+1}}$ is inexact, and this method fails to guarantee $\Pcal_{k+1}^l \in \Fcal$. Therefore, $\Mcal_k$ is not guaranteed to be correct. 

\subsection{Proposed Approach}

In the presence of odometry drift, since the transform $\T{B_k}{B_{k+1}}$ is uncertain, the method in~\eqref{eqn:basic_polytope} does not work. Extending this approach to uncertain transforms is also not straightforward, since in the H-representation, an uncertain perturbation to a half-space does not result in a new half-space. Here, we propose a novel method that uses the V-representation of the polytope to circumvent this issue. In the V-representation, the polytope is the convex-hull of a set of vertices. Denote the set of vertices by
\eqn{
        \Vcal_{i} = \{ v_{i, j} \}_{j=1}^{m_i} \subset \R^3,
}
where $v_{i,j} \in \R^3$ is the $j$-th vertex on the $i$-th face of a polytope.

We will use the V-representation to compute a new (deflated) polytope $\Pcal_{k+1}$ from $\Pcal_k$. The algorithm is described by the next Lemma and Theorem.

\begin{lemma}
        \label{lemma:polytope_separating}
        Suppose $\T{B_k}{B_{k+1}} \sim \Ncal(\estT{B_k}{B_{k+1}}, \Sigma_k)$, where 
        \eqn{
                \estT{B_k}{B_{k+1}} = \bmat{R & t \\ 0 & 1}.
        }
        Consider a polytope $\Pcal_k$ that is obstacle free, 
        \eqn{
                \Pcal_k = \{ p \in \R^3 : A_k p \leq b_k \}
        }
        where $A_k \in \R^{N \times 3}$, $b_k \in \R^N$. 
        Denote the $i$-th row as $a_{k, i} \in \R^3$.
        For each vertex $v_{i,j} \in \Vcal_i(\Pcal_k)$ on the $i$-th face of the polytope, define 
        \eqn{
                J_{i, j} = \bmat{ R & -R [v_{i, j}]_\times}, \quad 
                \Sigma_{i, j} = \kappa J_{i, j} \Sigma_k J_{i, j}^T, 
        }
        as in~\Cref{assumption:p}. Let each element of $\rho\in \R^N$ be 
        \eqn{
                \rho_i = \max_{j \in \{1, ..., m_i\}} \sqrt {a_{k, i}^T \Sigma_{i, j} a_{k, i}}
        }
        Define a new polytope as
        \seqn[eqn:shrunk_polytope]{
                \Pcal_{k+1} &= \{ p \in \R^3 : A_{k+1} p \leq b_{k+1} \},\\
                A_{k+1} &= A_k R^T,\\
                b_{k+1} &= b_{k} + A_k R^T t - \rho. \label{eqn:minus_rho}
        }
        Given~\Cref{assumption:p}, $\Pcal_k \in \Fcal\inframe{B_k} \implies \Pcal_{k+1} \in \Fcal \inframe{B_{k+1}}$, i.e., if $\Pcal_k$ is obstacle-free, so is $\Pcal_{k+1}$. 
\end{lemma}
\begin{IEEEproof}[Proof Sketch]
[See~\Cref{appendix:proofs:polytope_separating} for the full proof.] It suffices to show that any obstacle potentially on the boundary of $\Pcal_k$ will not be in $\Pcal_{k+1}$ after the rigid transform. To do so, we  consider a potential obstacle on the $i$-th face of the polytope, and compute the ellipsoid the obstacle could be in after the transform. We compute the tangent plane of the ellipsoid normal to the $i$-th hyperplane, and compute the minimum shift necessary such that the shifted hyperplane does not contain the ellipsoid. We use the convexity of the polytope to show that the necessary shift on the $i$-th hyperplane is $\rho_i$, the maximum of the shifts necessary at each of the vertices  on the $i$-th hyperplane of the polytope. This deflaion, when applied to each hyerplane of the polytope, guarantees that  $\Pcal_{k+1}$ does not contain the obstacle points.  
 \end{IEEEproof}

Finally, we can construct the main theorem. 

\begin{theorem}
        Suppose the transform between frame is $\T{B_k}{B_{k+1}} \sim \Ncal(\estT{B_k}{B_{k+1}}, \Sigma_k)$. Given the $k$-th map is defined as in~\eqref{eqn:Mcal_k}, define the $(k+1)$-th map as 
        \eqn{
                \Scal_{k+1}\inframe{B_{k+1}} = \bigcup_{l=1}^N \Pcal_{k+1}^l
        }
        where each polytope is defined using \Cref{lemma:polytope_separating}. Then, given \Cref{assumption:p}, 
        \eqn{
                \Scal_k \subset \Fcal \implies \Scal_{k+1} \subset \Fcal,
        }
        that is, if $\Mcal_k$ is correct by \Cref{defintion:correct_map}, the updated map $\Mcal_{k+1}$ will also be correct.
\end{theorem}
\begin{IEEEproof}
        Directly apply~\Cref{lemma:polytope_separating} to each polytope in $\Scal_k$. 
\end{IEEEproof}

In words, the theorem shows that when each polytope in the map $\Mcal_k$ is shrunk using ~\Cref{lemma:polytope_separating}, the new safe region $\Scal_{k+1}$ also remains certifiably obstacle-free. Once a given polytope has shrunk to zero volume, it can be forgotten entirely. Recall that as new camera frames are received, new polytopes can be constructed to define the free space in the operating environment and added to the set $\Scal_{k+1}$. We empirically study how quickly an environment deflates in~\Cref{tab:free_volume} and in~\Cref{appendix:collab_space}. Naturally, if the odometry covariance is smaller, the deflation rate is smaller~\Cref{appendix:effect_of_covariance}. 

\begin{remark}
        Compare \eqref{eqn:basic_polytope} with \eqref{eqn:shrunk_polytope}. The two are identical except for the $-\rho$ vector in~\eqref{eqn:minus_rho}. Each element $\rho_i\geq 0$, and therefore, this represents a shrinking operation. The net effect is that we transform the polytope by the estimated transform, but then shrink the polytope based on the odometry error covariance. Notice that this shrinking operation is tight: since there could exist an obstacle on the face of the polytope (indeed this is how they are constructed), the shrinking factor is the smallest allowable factor, by construction.  
\end{remark}
\begin{remark}
        In implementation, notice that one needs to compute $\Vcal_i(\Pcal_k)$, the set of vertices, and then update the polyhedron by~\eqref{eqn:minus_rho}.  Although this operation scales exponentially with the number of faces~\cite{fukuda1995double}, efficient implementations exist, especially for 3D polytopes~\cite{legat2023polyhedral}. Empirically, we observe each polytope has on the order of 10-20 faces when using~\cite{liu2017planning}, and can be handled in real-time. 
        
\end{remark}

\section{Approach 2: Certified \acp{ESDF}}
\label{section:certified_sdf}

\subsection{Background}

\edit{The \acf{ESDF} is defined as the function $d: \R^3 \to \R$,
\eqn{
\label{eqn:true_esdf}
d(p) = \begin{cases}
    \operatorname{dist}(p, \partial \Ocal), & \text{if } p \not \in \Ocal\\
    -\operatorname{dist}(p, \partial \Ocal),    & \text{if } p \in \Ocal
\end{cases}
}
where $\partial \Ocal \subset \R^3$ is the boundary of the obstacles. The $\operatorname{dist}$ measures the minimum distance of a point to a set, i.e., $\operatorname{dist}(p, \partial \Ocal) = \min_{o \in \partial \Ocal} \norm{p - o}$. Thus, for any point in free-space,\footnote{We use~\eqref{eqn:esdf} instead of~\eqref{eqn:true_esdf} for the remainder of the section for brevity. The points with $d(p) < 0$ will be removed from memory.} the \ac{ESDF} is given by
\eqn{
\label{eqn:esdf}
d(p) = \min_{o \in \Ocal} \norm{ o - p},
}
A 2D slice of the \ac{ESDF} is depicted in~\Cref{fig:mapping_outputs}d. 
}

To evaluate~\eqref{eqn:esdf},  $o$ and $p$ must be expressed in a common frame, commonly referred to as the mapping frame. Since this is done in the mapping frame, it is denoted as the function $d_M : \R^3 \to \R$.
The claimed-safe region $\Scal_k$ is therefore
\eqn{
\Scal_k = \{ p \in \R^3 : d_M(p) \geq 0 \}
}

For safety-critical path planning and control, we need the \ac{ESDF} at points relative to the body-fixed frame. 
The common approach is to assume the odometry estimate is exact, and determine $d(p\inframe{B_k})$ by expressing it in the map frame and evaluating $d_M$:
\eqn{
        d(p\inframe{B_k}) \approx  d_M( \estT{B_k}{M} \cdot p\inframe{B_k} ) \label{eqn:incorrect_esdf}
}
However, since the estimate $\estT{B_k}{M}$ is inexact, this method can lead to over- or under-estimates. Overestimated distances are unsafe since they could lead to collisions. 

\subsection{Proposed Approach}

The goal is to construct an \ac{ESDF} that is safe, i.e., underestimates the distance to obstacles. Using~\Cref{defintion:correct_map}, a \emph{Certified-ESDF} is defined as
\begin{definition}
\label{def:cesdf}
    Let the obstacle set be $\Ocal \subset \R^3$, assumed static in frame $I$. Let the \ac{ESDF} of $\Ocal$ be $d: \R^3 \to \R$. A \emph{Certified-ESDF} (C-ESDF) at timestep $k$ is a function $d_M^k: \R^3 \to \R$, such that for all points $p\inframe{B_k} \in \R^3$,
\eqn{
        d(p\inframe{B_k}) &\geq d_M^k( \estT{B_k}{M} \cdot p \inframe{B_k} ) \label{eqn:correct_esdf}
}
where $\estT{B_k}{M} \in \SE(3)$ is the estimated rototranslation between $B_k$ and $M$.
\end{definition}

Comparing \eqref{eqn:incorrect_esdf} with \eqref{eqn:correct_esdf}, the goal of certification is to change the $\approx$~into~$\geq$. 
That is, a Certified-ESDF is one where for any body-fixed point $p\inframe{B_k}$, if the point is expressed in the mapping frame \emph{using the estimated rototranslation}, we have \emph{underestimated} the distance to the nearest obstacle:
\eqn{
\underbrace{d(p\inframe{B_k}) = \min_{o \in \Ocal} \norm{ p\inframe{B_k} - o\inframe{B_k}}}_{\text{true ESDF}} &\geq \underbrace{d_M(\estT{B_k}{M} \cdot p\inframe{B_k})}_{\text{estimated ESDF}}.
}

To accomplish this, we propose a strategy of deflating the \ac{ESDF}. We derive a recursive guarantee to ensure the \ac{ESDF} remains certified for all $k$. 

\begin{theorem}
\label{theorem:esdf_theorem}
        Suppose at timestep $k \in \naturals$, the \ac{ESDF} $d_M^k: \R^3 \to \R$ is a Certified-ESDF. 
        Let the rototranslation between frames be $\T{B_{k+1}}{B_{k}} \sim \Ncal(\estT{B_{k+1}}{B_{k}}, \Sigma_k)$. 
        Let the $d_M^{k+1}: \R^3 \to \R$ be defined by 
        \eqn{
                d_M^{k+1}(p\inframe{M}) = d_M^{k}(p\inframe{M}) - \sqrt{\lambda_{\max}(\Sigma_p)} \label{eqn:lamda_max}
        }
        for all $p\inframe{M} \in \R^3$, where
        \seqn{
                \estT{B_{k+1}}{B_k} &= \bmat{R & t \\ 0 & 1},\\
                J &= \bmat{ R & -R [\estT{M}{B_{k+1}} \cdot p\inframe{M}]_\times},\\
                \Sigma_p &= \kappa J \Sigma_k J^T.
        }
        and $\kappa > 0$ is as defined in~\Cref{assumption:p}.
        Given~\Cref{assumption:p}, $d_M^{k+1}$ is also a Certified-ESDF at timestep $k+1$. 
\end{theorem}

\begin{IEEEproof}[Proof Sketch]
[See~\Cref{appendix:proof:esdf_theorem} for the full proof.] Consider any point $p\inframe{B_{k+1}}$ and evaluate the potential positions it could correspond to in frame $B_k$. This is an ellipsoid as in~\Cref{assumption:p}, and therefore the \ac{ESDF} at $p\inframe{B_{k+1}}$ must be the minimum of all of the \ac{ESDF} values for the corresponding points in the ellipsoid. Since, by definition, the Lipschitz constant of an \ac{ESDF} is one, this minimum \ac{ESDF} can be lower bounded by the \ac{ESDF} at the center minus the radius of the smallest sphere containing the ellipsoid. We use the eigenvalues of the ellipsoid to compute the radius of sphere, arriving at the expression. 
\end{IEEEproof}

\begin{remark}
Notice that the correction is $-\sqrt{\lambda_{max}(\Sigma_p)}$ in~\eqref{eqn:lamda_max} (different for each $p$). As with the certified \acp{SFC}, this is a deflation operation that decreases the estimated distance to an obstacle. 
\end{remark}

\begin{remark}
    The implementation of this deflation operation is remarkably simple and easily parallelized on a GPU.  In our implementations, we added an additional deflation integrator to the code in~\cite{nvblox}. At each frame, when the relative odometry with covariance is received, we can compute the deflation at each voxel in parallel using~\eqref{eqn:lamda_max}. 
\end{remark}

\section{Safe Navigation with Certified Maps}
\label{section:safe_navigation}

Here we summarize the key ideas presented in this paper, and suggest strategies to achieve safe navigation. 

A fundamental principle of our approach is ensuring that maps remain correct with respect to the body-fixed frame. To achieve this, we deflate the safe regions of the map based on the incremental odometry error at each timestep. The required deflation has an analytic expression.

Our implementation is as follows. When the $(k+1)$-th camera frame is received from the sensor, we compute the odometry estimate, and its relative covariance. Next, we apply the deflation step using the  proposed algorithms. Finally, we incorporate new safe regions identified by the depth image to assimilate new information while discarding regions that can no longer be certifiably correct. 


One can also maintain both the baseline and certified maps in memory simultaneously. While the memory usage increases, the certified maps tend to be smaller than the full map, maintaining both maps offers significant advantages.  In particular, our certified mapping methods can integrate naturally with existing safety filtering methods like~\cite{agrawal2024gatekeeper, tordesillas2019faster}. These methods generate nominal trajectories to achieve mission objectives, but use a backup trajectory to ensure that the robot can safely stop based on the currently available information. In our framework, one can use the baseline map for nominal trajectory planning, but use the certified map for collision and safety checks. This combination enables agile motion while strictly guaranteeing safety.

\section{Simulations}
\label{section:simulations}
\begin{figure*}
    \centering
        \includegraphics[width=0.95\linewidth]{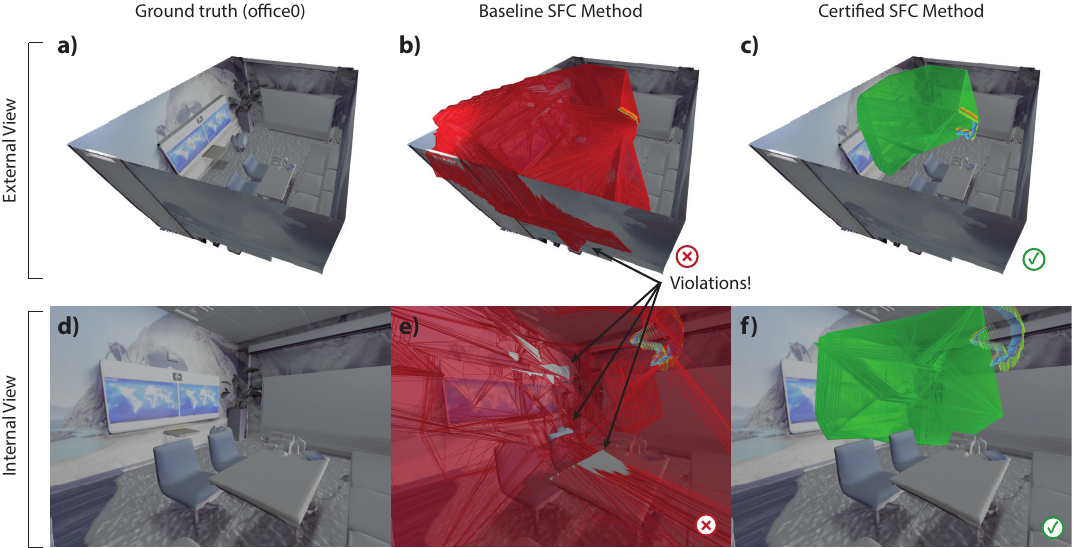}
    \caption{Visualization of a snapshot of the \texttt{office0} environment mapped using the baseline and certified~\ac{SFC} methods. (a, d) shows the \texttt{office0} environment, while (b, e) and (c, f) show the respective $\Scal$ sets at the 500-th timestep from an external and an internal view.  The baseline map claims a larger volume to be safe compared to the certified method (red volume is larger than green volume). However, we can also see numerous regions where the red region intersects with the ground truth mesh, indicating that the claimed safe region contains obstacle points. In the certified method, we see no violations. }
    \label{fig:sfc_summary}
\end{figure*}

\begin{figure*}
    \centering
        \includegraphics[width=0.95\linewidth]{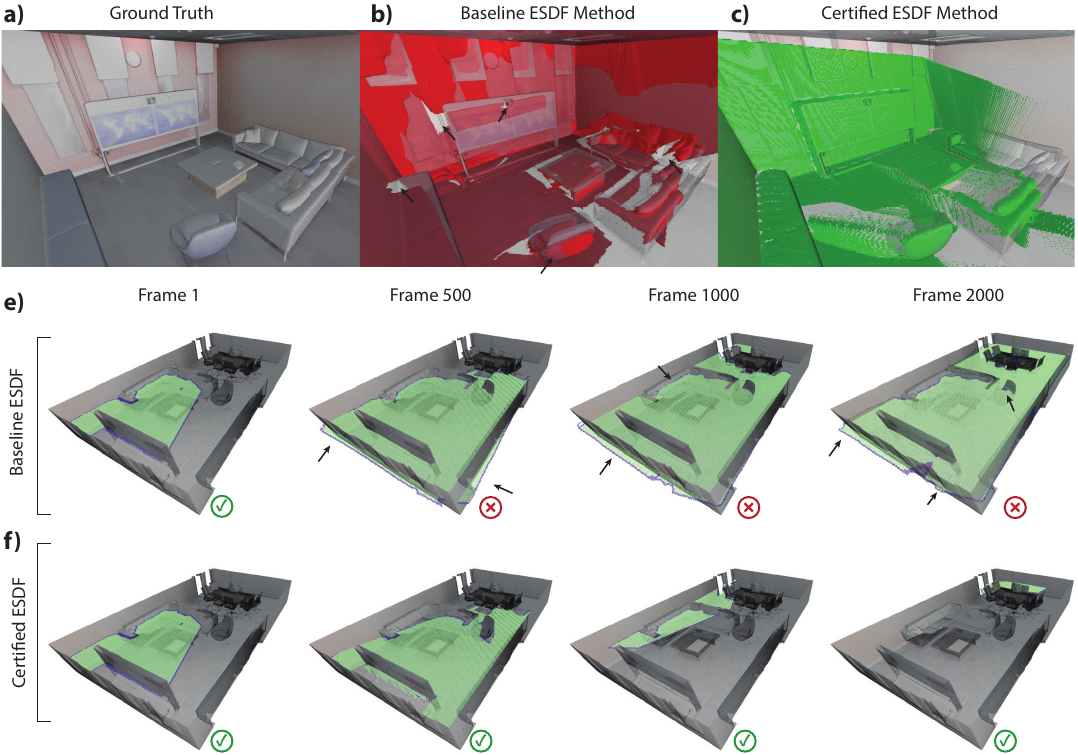}
    \caption{Visualization of the maps generated using the baseline and certified \ac{ESDF} methods on the \texttt{office3} environment. In (a) we see the ground-truth mesh. In (b) and (c) we can see the internal view after 500 timesteps. As in~\Cref{fig:sfc_summary}, although the baseline method maps a larger volume (red mesh is larger than green mesh), it also contains many violations. In (e) and (f) we see a slice of the \ac{ESDF} over time. The green region indicates the $\Scal$ set at the respective times. The small black arrows point to various violations in the baseline method, while in the certified methods we see no violations.} 
    \label{fig:sdf_summary}
\end{figure*}

We present results on the accuracy and correctness of both approaches for certified mapping presented above. As a reminder, the goal is to demonstrate that despite odometry drift, the region reported by our algorithms to be a part of the free space is indeed obstacle-free. First, we evaluate the performance of both the Certified~\acp{SFC} and the Certified~\acp{ESDF} methods on the Replica dataset (described below) and compare it to various baselines. Second, we have run hardware experiments with a rover, and show that by considering the certification bound the rover can avoid collisions. Additional results are reported in~\Cref{appendix:additional} and~\Cref{appendix:effect_of_covariance}.

\subsection*{Evaluation Method}
We evaluated the performance of our implementations on the Replica dataset~\cite{replica19arxiv}, with ground-truth trajectories generated as in~\cite{Zhu2022CVPR}. From the ground-truth trajectory the RGBD image sequence was generated. We perturbed the trajectory to generate the estimated trajectory from a simulated odometry system as follows:
\eqn{
\estT{B_{k+1}}{B_{k}} = \T{B_{k+1}}{B_{k}} \Exp( \tau), \quad \tau \sim \Ncal(0, \Sigma)
}
where $\T{B_{k+1}}{B_{k}} \in \SE(3)$ is the transform between subsequent frames of the ground-truth trajectory of the camera and $\estT{B_{k+1}}{B_{k}} \in \SE(3)$ is the estimated transform between subsequent frames used in the mapping algorithms. We used $\Sigma \in \{ 10^{-5} I, 10^{-6} I\}$. Evaluating the Absolute Translation Error (ATE) as in~\cite{zhang2018tutorial}, the generated trajectories had between $1-3\%$ ATE, inline with the performance of state-of-the-art \ac{VIO} methods. Each trajectory has 2000~frames at 30~FPS. 

\subsection*{Baselines}

We compared our proposed certified approaches to the following mapping methodologies:
\begin{enumerate}
    \item [(A)] \emph{Baseline \ac{SFC}} - At each camera frame, the depth map is used to construct a pointcloud of obstacles within the current field of view. From this a convex polyhedron is extracted, and appended to a list of safe polyhedrons. The union of these polyhedrons is considered the safe flight region. We used the library~\cite{liu2017planning} to perform the convex decomposition. 
    \item [(B)] \emph{Heuristic \ac{SFC}} - This is the same algorithm as in (A), except that a time-based forgetting mechanism is introduced, as is common in robotic mapping implementations. In particular, we only keep the last 60~frames (2~seconds) of polyhedrons when constructing the safe flight region. 
    \item [(C)] \emph{Baseline \ac{ESDF}} - At each camera frame, the depth map is used to update the \ac{TSDF} of the environment. At regular intervals a wave propagation algorithm constructs/updates the \ac{ESDF} of the environment. Regions with positive \ac{ESDF} are considered part of the safe flight region. We used the library~\cite{nvblox} to construct the \ac{TSDF} and \ac{ESDF}.
    \item [(D)] \emph{Heuristic \ac{ESDF}} - This is the same algorithm as in (C), except that a distance-based forgetting mechanism is introduced. In particular, we forget any \ac{TSDF} and \ac{ESDF} voxels that are more than 3~m away from the camera.
\end{enumerate}

These are compared to the proposed certified methods:
\begin{enumerate}
    \item [(E)]  \emph{Certified \ac{SFC}} - This is the same algorithm as in (A), except that at each frame, each polytope is deflated as described in~\Cref{section:certified_sfc}.
    \item [(D)] \emph{Certified \ac{ESDF}} - This is the same algorithm as in (C), except that at each frame, the \ac{ESDF} is deflated as described in~\Cref{section:certified_sdf}.
\end{enumerate}

\subsection*{Metrics}

To evaluate the performance, we consider three metrics:
\begin{enumerate}
    \item [(I)] \emph{Violation Rate:} The violation rate measures the percentage of ground-truth mesh points that (incorrectly) lie within the claimed free space. The violation rate should be close to 0\%.
    \item [(II)] \emph{Maximum Violation Distance: } For any violating point we measure the maximum distance of the violation, i.e., how far into the claimed free space is an obstacle point. The violation distance should be close to 0~mm. If there are no violating points, the violating distance is 0~mm. 
    \item [(III)] \emph{Free-Space Volume:} This measures the total volume of the space that is claimed to be free.  The free-space volume should be as large as possible. 
\end{enumerate}


\subsection*{Results}

\begin{table*}[t]

\centering
\caption{Violation Rates. This table summarizes the fraction of violating ground-truth obstacle points for each environment and algorithm. This table shows results with $\Sigma = $1e-6$I$.} 
\label{tab:violation_rate}
\resizebox{0.95\textwidth}{!}
{%
\begin{tabular}{@{} c | S[table-format=4.4] S[table-format=4.4] S[table-format=4.4] S[table-format=4.4] S[table-format=4.4] S[table-format=4.4] S[table-format=4.4] S[table-format=4.4]  @{} }
\toprule
& \multicolumn{8}{c}{Violation Rates (\pct)}\\
Algorithm           & \texttt{office0}  & \texttt{office1}  & \texttt{office2}  & \texttt{office3}  & \texttt{office4}  & \texttt{room0}    & \texttt{room1}    & \texttt{room2}    \\
\midrule
Baseline SFC & 	18.60\pct          & 	12.76\pct & 	10.13\pct        & 	12.74\pct        &   	14.44\pct &        	10.74 \pct & 	19.17   \pct &       	6.85 \pct \\
Heuristic SFC & 	0.11\pct          & 	0.57\pct & 	0.09\pct      & 	0.10\pct         &   	0.27\pct &         	0.02 \pct & 	0.39 \pct &       	0.92 \pct \\
\rowcolor{gray!30} Certified SFC & 	0.0002\pct            & 	0.0047\pct & 	0.0008\pct      & 	0.0005\pct       &   	0.0014\pct &       	0.0002 \pct & 	0.0009 \pct &       	0.0012 \pct \\
Baseline ESDF & 	48.15\pct         & 	35.31\pct & 	51.51\pct        & 	54.66\pct        &   	48.35\pct &        	62.03 \pct & 	48.15   \pct &       	47.49 \pct \\
Heuristic ESDF & 	31.55\pct            & 	34.39\pct & 	7.63\pct         & 	4.66\pct         &   	10.08\pct &        	9.25 \pct & 	20.88    \pct &       	16.32 \pct \\
\rowcolor{gray!30} Certified ESDF & 	0.5443\pct           & 0.0610\pct & 0.0809\pct      & 0.0227\pct       &   0.0538\pct &     2.4259 \pct & 0.0149 \pct &  0.0519 \pct \\
\bottomrule
\end{tabular}%
}

\vspace{1em}

\caption{Maximum Violation Distance. This table summarizes the distance by which violating ground-truth obstacle points penetrate the estimated free space for each environment and algorithm. This table shows results with $\Sigma = $1e-6$I$.}
\label{tab:max_violation}
\resizebox{0.95\textwidth}{!}
{%
\begin{tabular}{@{} c | S[table-format=4.4] S[table-format=4.4] S[table-format=4.4] S[table-format=4.4] S[table-format=4.4] S[table-format=4.4] S[table-format=4.4] S[table-format=4.4]  @{} }
\toprule
& \multicolumn{8}{c}{Maximum Violation Distance (mm)}\\
Algorithm           & \texttt{office0}  & \texttt{office1}  & \texttt{office2}  & \texttt{office3}  & \texttt{office4}  & \texttt{room0}    & \texttt{room1}    & \texttt{room2}    \\
\midrule
Baseline SFC &    	102.7 & 95.3  & 159.7 & 177.6 & 125.5 & 117.1 & 191.4 & 85.0  \\
Heuristic SFC &    22.1  & 14.5  & 18.4  & 11.6  & 8.9   & 11.0  & 14.2  & 12.8  \\
\rowcolor{gray!30} Certified SFC &    0.0   & 0.9   & 0.4   & 0.9   & 1.7   & 0.9   & 0.7   & 0.7   \\
Baseline ESDF &    604.3 & 406.9 & 520.0 & 671.1 & 636.9 & 990.8 & 604.6 & 594.0 \\
Heuristic ESDF &   563.6 & 379.5 & 311.8 & 429.4 & 366.6 & 428.5 & 384.7 & 435.4 \\
\rowcolor{gray!30} Certified ESDF &   109.5 & 82.5  & 141.4 & 100.0 & 66.3  & 120.0 & 100.0 & 82.5 \\
\bottomrule
\end{tabular}%
}
\vspace{1em}
\caption{Estimated Free Space Volume. This table summarizes the volume of the estimated free space at the end of the simulation for each environment and algorithm. This table shows results with $\Sigma = $1e-6$I$.}
\label{tab:free_volume}
\resizebox{0.95\textwidth}{!}
{%
\begin{tabular}{@{} c | S[table-format=4.4] S[table-format=4.4] S[table-format=4.4] S[table-format=4.4] S[table-format=4.4] S[table-format=4.4] S[table-format=4.4] S[table-format=4.4]  @{} }
\toprule
& \multicolumn{8}{c}{Estimated Free Space Volume (m$^3$)}\\
Algorithm           & \texttt{office0}  & \texttt{office1}  & \texttt{office2}  & \texttt{office3}  & \texttt{office4}  & \texttt{room0}    & \texttt{room1}    & \texttt{room2}    \\
\midrule
Baseline SFC &    34.8 & 17.6 & 40.8 & 56.6  & 63.3 & 53.0  & 38.7 & 29.4 \\	
Heuristic SFC &   6.7  & 3.6  & 4.3  & 4.6   & 15.7 & 12.3  & 6.9  & 7.5  \\
\rowcolor{gray!30} Certified SFC &   5.7  & 2.6  & 3.6  & 3.0   & 12.5 & 9.1   & 5.8  & 4.4  \\
Baseline ESDF &   46.1 & 23.2 & 77.5 & 110.9 & 99.7 & 105.4 & 53.8 & 63.6 \\
Heuristic ESDF &  39.5 & 23.0 & 31.3 & 42.0  & 51.5 & 28.6  & 34.5 & 38.7 \\
\rowcolor{gray!30} Certified ESDF &  10.7 & 3.8  & 6.2  & 5.0   & 14.3 & 31.5  & 6.6  & 4.5  \\
\bottomrule
\end{tabular}%
}
\end{table*}

Tables~\ref{tab:violation_rate}, \ref{tab:max_violation}, and \ref{tab:free_volume} summarize the results from the simulations. \Cref{fig:sfc_summary} and \Cref{fig:sdf_summary} visualize the results and qualitatively show the behavior of the proposed methods. 

\Cref{fig:sfc_summary} visualizes one of the runs from the \texttt{office0} environment. Figures (a, d) shows the ground-truth mesh of the environment from two different views. In (b, e) we see the safe flight polytopes in the baseline method visualized as the red region. One can see that the red region clearly intersects with the ground-truth mesh, and each intersection represents a violation. The violations are particularly noticeable for regions that were mapped further in the past, and from non-convex and thin obstacles like the chair or table surfaces. In contrast, in (c, f) we see the safe flight polytope from the proposed certified algorithms, drawn as the green region. We can see that the green region is smaller than the red polytope, but it also contains no violating points (see also \Cref{tab:max_violation} and~\Cref{tab:free_volume}). Effectively, we can see that due to the odometry drift, the algorithm cannot be confident about the exact location of, for example, the chair and the desk, and therefore these regions were removed from the map. Although the volume of free space is smaller, the map is guaranteed to be correct.  

From~\Cref{tab:violation_rate} we can observe that both certification methods significantly reduce the number of violations. In the baseline methods, the violation rates are between 6 and 60\%, while in the certified methods, the violation rates are between 0-3\%. Note, we cannot expect the certified methods to have exactly zero violations, since we are using the truncated noise model for odometry. Nonetheless, empirical performance of the certified methods still shows that the proposed methods can effectively avoid classifying obstacle regions as free. 

Furthermore, we can see that although the heuristic forgetting methods can also reduce the number of violations, the level of reduction is hard to control. Since the forgetting factor is tuned heuristically and independently of the true noise level in the system, it can sometimes lead to good rejection of obstacles (as in the \ac{SFC} method) or poor rejection of obstacles (as in the \ac{ESDF}).  

From~\Cref{tab:max_violation} we observe that the maximum distance a violating point intersects the map is also reduced using the certified methods. We see that the maximum violation is sub-millimeter for the \ac{SFC} methods, demonstrating a reduction of 2 orders of magnitude compared to the baseline. In the \ac{ESDF} approaches, we still see a significant reduction in the maximum violation distance (about an order of magnitude reduction), although there are some violations on the order of 100~mm. This seems to be a limitation of the \ac{ESDF} approach, since the \acp{ESDF}  are represented using discrete voxels computationally. We chose a voxel size of 20~mm, and therefore the violations are on the order of 1-5~voxels of error.\footnote{Finer grid resolution can help, but will increase the computational and memory requirements. As a sense of scale, each environment is on the order of $6\times 6 \times 3$~m, and therefore has approximately $300 \times 300 \times 150$ voxels. See~\Cref{tab:bounding_boxes} for additional details. }

The source of this larger error is likely the dataset itself. We have checked which voxels are causing these large errors, and it seems to be the voxels that are close to non-manifold surfaces in the Replica dataset, for instance near the leaves of plants, or around table/chair legs, which are thin and long. Near these surfaces, the raw data is inconsistent, and we suspect that it leads to higher error rates than expected. 

Finally, we can see that due to the certification the volume of the estimated free space is lower for the certified methods than it is for the heuristic or baseline methods (\Cref{tab:free_volume}). However, since the violation rate of the uncertified methods is significant, the free space cannot be trusted for path planning around obstacles. Despite the smaller volume of free space, the certified methods allow the full region to be trusted when used in planning (\Cref{appendix:collab_space}).

Comparing the \ac{SFC} and \ac{ESDF} methods, in the results presented the \ac{SFC} methods seem superior, since they have fewer violations, and the violating points violate the free space by a smaller distance. However this does come at the expense of expressiveness and computational cost. The \ac{SFC} methods require the use of unions of convex polytopes to represent the free space, and in cluttered environments can sometimes lead to very small volumes of free space. The \ac{ESDF} implementations are also more mature, with implementations like~\cite{nvblox} allowing for efficient use of a GPU, which allows the \ac{ESDF} to be computed more efficiently than the \ac{SFC}.

\section{Rover Experiments}
\label{section:experiments}

In this section we demonstrate the utility of the proposed certified mapping frameworks in ensuring a robot can safely navigate an environment. We demonstrate that when a rover is tasked to navigate through an environment, and in particular reverse blindly into a region it previously mapped, the accumulated odometry error can lead to the rover colliding with previous mapped obstacles. Instead, by using the proposed methods, the rover will avoid traversing into regions that it can no longer certify are obstacle-free. 

\subsection*{Experimental Setup}
\begin{figure}[t]
    \centering
    \includegraphics[width=\linewidth]{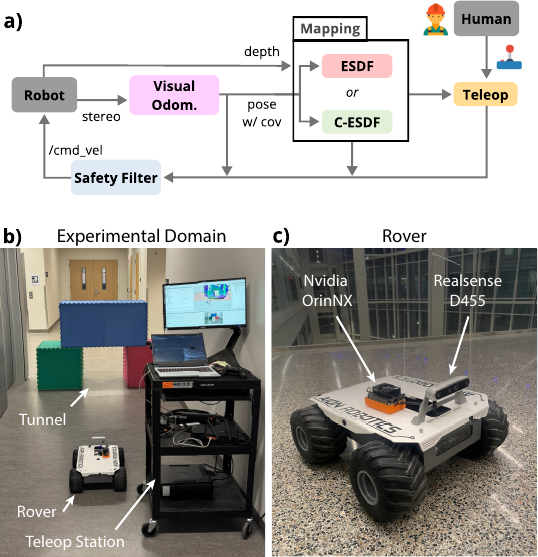}
    \caption{Rover Experimental Setup. (a) Block diagram. The human is teleoperating the rover using only the \ac{FPV} feed and the reconstructed obstacle map computed and streamed in real-time. The map is also used onboard the robot to stop the robot if it violates safety constraints. The safety filter can either use the baseline \ac{ESDF} or the Certified~\ac{ESDF}. (b) Picture of the testing environment. The robot drives through the tunnel, mapping it as it passes through. After exploring the corridors, the rover tries to return through the tunnel in reverse, without remapping the tunnel. (c)~shows the rover in more detail. The AION R1 UGV has been modified, with all sensing on Intel Realsense D455, and all compute on the Nvidia OrinNX 16GB.}
    \label{fig:exp_setup}
\end{figure}

\begin{figure*}[t]
    \centering
    \includegraphics[width=\linewidth]{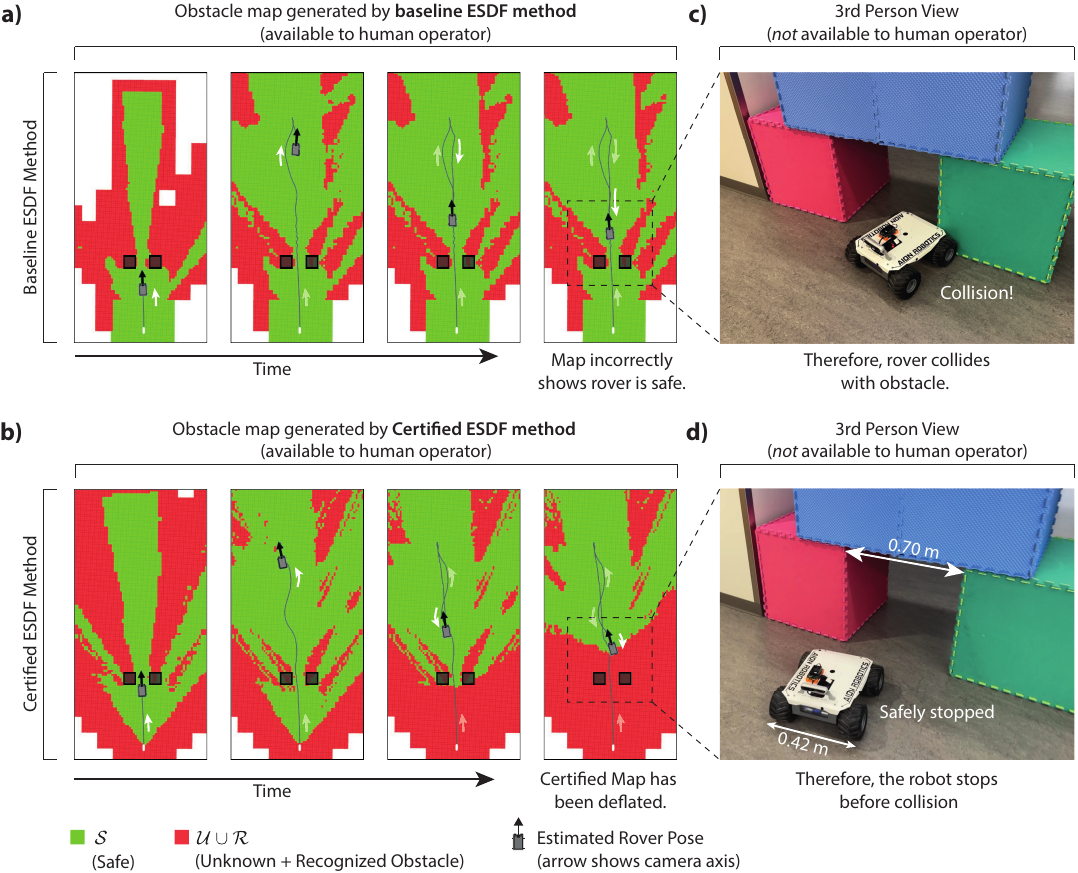}
    \caption{Rover Experimental Results. (a, b)~shows snapshots of the reconstructed obstacle map and the estimated rover pose with (a)~the baseline method and (b)~the certified method. This is the view presented to the human teleoperating the robot. Note, two small black boxes are drawn in each frame (in post) to indicate to the reader the location of the red and green boxes during the experiment. These were not visible to the human operator during the experiments. (c, d)~show the final state of the robots at the end of the trajectory. In (c), the baseline method the robot has crashed with the green obstacle, although looking at the last panel of (a), we can see that the robot thinks it is in the middle of the tunnel in the free space. In (d), we see the robot stopped 15~cm before crashing with the red obstacle, and this is because the map has been deflated sufficiently that the safety filter prevents the robot from continuing backwards. Notice between the second, third and fourth frames in~(b) the green regions near the bottom change into red regions, indicating the Certified~\ac{ESDF} cannot certify that the red region is obstacle-free.}
    \label{fig:rover_experiments}
\end{figure*}

A block diagram of the experimental setup is shown in~\Cref{fig:exp_setup}a). We use a ground rover, the AION R1 UGV equipped with an Intel Realsense D455 camera. All perception, planning, and control is executed on the onboard computer, an Nvidia Orin NX 16GB. The Realsense camera sends stereo infrared images to the Orin NX at 30FPS. A state-of-the-art visual slam algorithm (Nvidia IsaacROS Visual SLAM) is used to compute the odometry estimate. The Realsense camera also produces a depth image, which is sent to the obstacle mapping library (an adapted version of Nvidia IsaacRos NvBlox) which constructs an \ac{ESDF} of the environment in real-time. All parameters and code  is available at [redacted].

A human operator uses a joystick to send desired linear and angular velocities to the robot. Using the constructed \ac{ESDF}, a safety filter forward propagates the robot's state under a desired command a short (0.5~s) horizon into the future and checks whether the trajectory lies strictly within $\Scal_k$. If so, the command is sent to the robots' motor controllers. If not, the safety filter zeros the linear command, and sends a reduced angular speed command. This allows the robot to continue to spin to acquire new information about the environment, without physically moving and potentially colliding with the obstacles. The safety filter was tuned offline to ensure that in the absence of odometry drift, the robot stops within 15~cm of the obstacle both when driving forwards or backwards. 

To compute the certified-correct map, we use the techniques of~\Cref{section:certified_sdf} to compute the certified~\ac{ESDF} representing the local geometry. To correctly deflate the \ac{ESDF}, we require the odometry estimate, and the covariance of the incremental transform between successive camera frames, i.e., of $\estT{B_{k-1}}{B_k}$. 

To the best of the author's knowledge however, this information is not reported by any state-of-the-art odometry/pose estimation algorithms. Most algorithms (including Nvidia's vSLAM) only report the covariance of the odometry estimate between the initial frame and the current frame, i.e., of $\estT{B_0}{B_k}$.  In~\cite{mangelson2020characterizing} the authors computed the covariance of relative poses after solving a pose-graph optimization problem by using the Jacobian of the local solution (see \cite[Section IX.B]{mangelson2020characterizing} for details). However this only allows one to find the covariance of relative transforms between keyframes, and does not allow one to find the relative transform between successive camera frames. \edit{Note, \cite{maken2021stein} reports the error covariance for frame-to-frame pointcloud matching, and could be integrated into the experiments below. However the accuracy of the pointcloud reported by the \ac{RGBD} camera must also be considered~\cite{nguyen2012modeling}.}

Here, we use the following method to estimate the covariance between relative frames. VSLAM reports the odometry estimates $\estT{B_0}{B_k}$, $\estT{B_0}{B_{k+1}}$, and the associated covariances $\Sigma_{B_0}^{B_k}$, $\Sigma_{B_0}^{B_{k+1}}$. Assuming $\T{B_0}{B_k}$ and $\T{B_0}{B_{k+1}}$ are highly correlated since they are successive frames, we can define a correlation coefficient $\rho \in [-1, 1]$ (we use $\rho = 0.99$) between these camera frames. We can then estimate the covariance of the relative transform $\Sigma_{B_k}^{B_{k+1}}$ along the lines of~\cite{mangelson2020characterizing}. The analysis is presented in~\Cref{appendix:extracting_relative_covariance}.

\subsection*{Experimental Results}

\Cref{fig:rover_experiments} summarizes the results of the rover experiments, with additional trials available in the supplementary video, all demonstrating similar outcomes. 

The human operator's task was to navigate the rover without line-of-sight through a narrow tunnel, explore and map the environment, and return to the starting location by reversing through the tunnel. The rover was intentionally reversed through the tunnel to avoid re-mapping the obstacle geometry, forcing it to rely on its previously constructed maps. 

Snapshots in~\Cref{fig:rover_experiments}a show the baseline method. Initially, the tunnel and the surrounding corridors are mapped accurately. As the operator tries to reverse through the tunnel the final snapshot suggests that the rover is well aligned with the tunnel and is within the green region $\Scal$. However, despite this seemingly safe alignment, the rover collided with an obstacle~\Cref{fig:rover_experiments}c, a failure in the baseline mapping approach. 

In contrast, our proposed method deflates the safe regions in response to the odometry drift. In~\Cref{fig:rover_experiments}b, the map initially classifies a large region as safe (green). However, as rover reverses to the tunnel, the deflation has caused parts of the map to turn red, indicating that these areas can no longer be certified to be obstacle free.  Indeed, when the rover reaches the boundary between red and green regions, the safety filter prevents further motion, successfully preventing collision. The same behavior was consistently observed across multiple trials.


\subsection*{Larger Scale Experiments}
\label{appendix:collab_space}

\begin{figure}[tb]
    \centering
    \includegraphics[width=\linewidth]{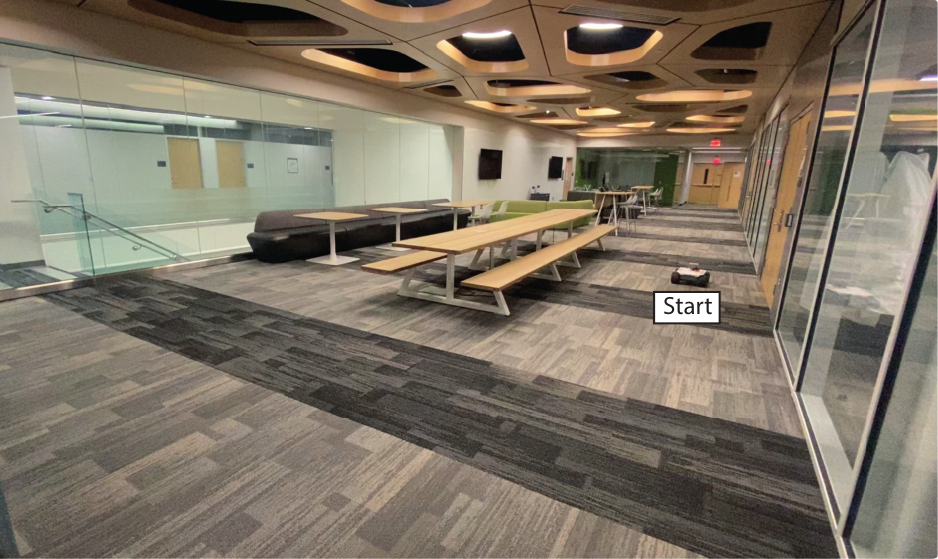}
    \caption{Experimental domain used in~\Cref{fig:decay_maps}.}
    \label{fig:collab_space}
\end{figure}

In this section, we show qualitatively and quantitatively the volume of free space usable by a robotic system. The rover was operated in a room approximately $40 \times 20$~m large drawn in~\Cref{fig:collab_space}. Starting in the middle, the robot was teleoperated to explore and map the room. The robot has a horizontal field of view of 75$^\circ$, and a maximum depth integration distance of 8~m. This means that from the depth image, the maximum distance that NvBlox will mark as free or safe is 8~m from the camera origin. Thus, in these experiments, the heuristic method also uses a forgetting radius of 8~m.

\begin{figure}[ht]
    \centering
    \includegraphics[width=\linewidth]{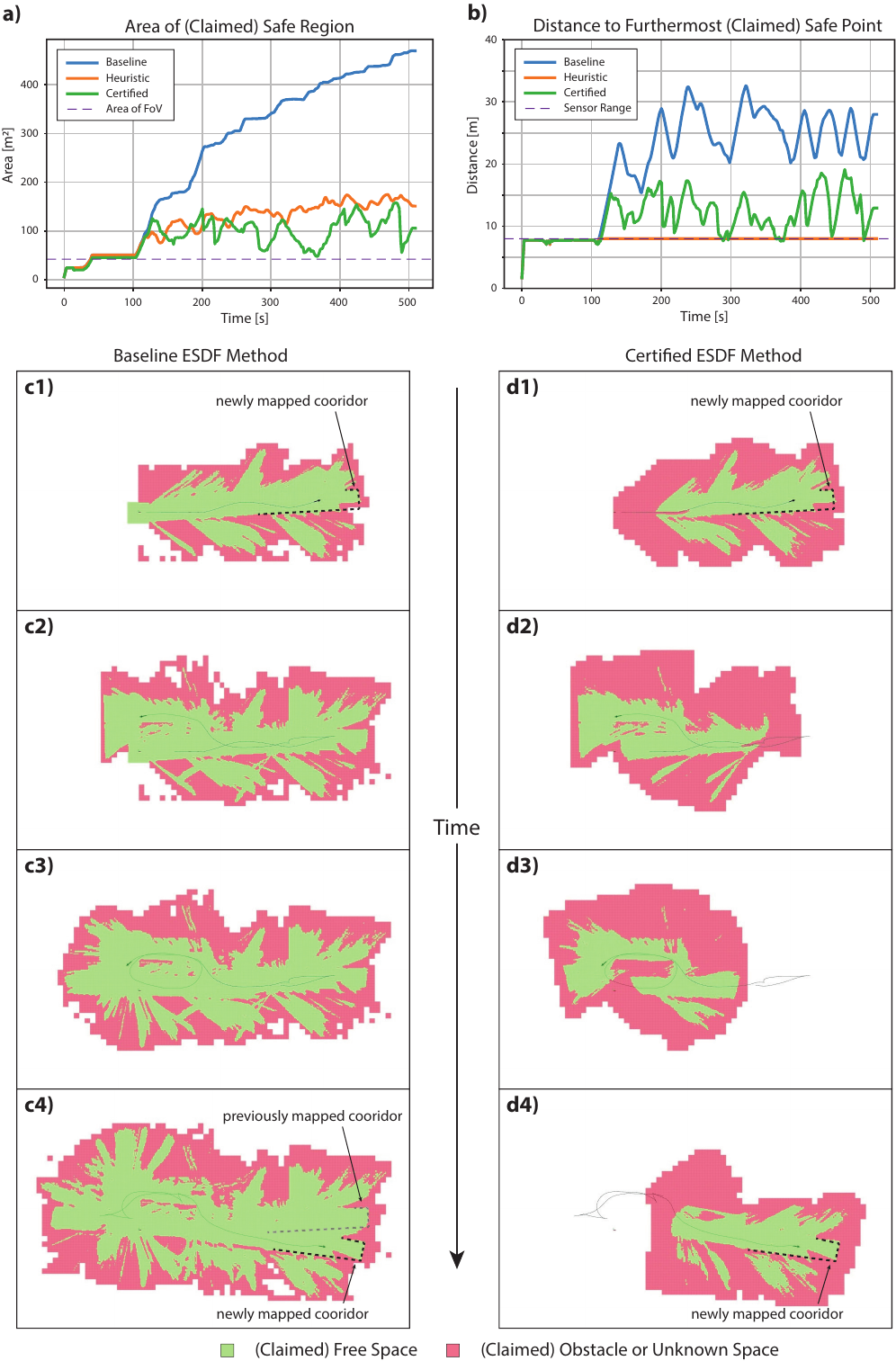}
    \caption{Quantitative and qualitative analysis of the effect of the deflation on the volume of the certified free space. (a)~Compares the area of the claimed safe region on a 2D slice of the \ac{ESDF} extracted at the robot height. As a reference, the area of the \ac{FOV} of the camera is also drawn (black dashed line). (b)~Compares the distance of the furthermost (claimed) free voxel from the robot position. As a reference, the maximum depth of the depth sensor (8~m) is indicated (black dashed line). In~(c1-c4) we see snapshots of the map generated by the Baseline \ac{ESDF} method, and in~(d1-d4) we see the corresponding snapshots from the Certified \ac{ESDF} method. The supplementary video animates the map slices.}
    \label{fig:decay_maps}
\end{figure}

A quantitative comparison of the algorithms is presented in~\Cref{fig:decay_maps}a, b. In~(a) we can see the area of the claimed safe region by each of the three methods. Although the claimed free region is largest for the baseline method, the map is erroneous. The certified and heuristic methods have similar free area, although the heuristic method is also often incorrect. 

In~\Cref{fig:decay_maps}b, we show the distance to the furthermost safe point from the robot position. This gives an indication of extent of the map that would be free if it were not for the obstacles in the environment. Here, we can see that compared to the maximum integration distance of 8~m, the certified method has its furthermost safe voxel approximately 12~m away, and upto 18~m away. In contrast, the heuristic method is clipped at 8~m. The evolution of the maps in time is clearer in the accompanying video, where the \ac{FPV} and third person view of the robot are also drawn.

Slices of the \ac{ESDF} and the Certified~\ac{ESDF} are shown in~\Cref{fig:decay_maps}c, d. The robot's trajectory is also drawn. Compare~\Cref{fig:decay_maps}c1-c4. We can see that the map drifts significantly - in (c1) we use a gray dashed line to highlight the end of the corridor as mapped at that time. In (c4), we draw the corridor mapped in (c1) as well as the newly mapped corridor, and we can see a significant shift in the map. In (d1-d4) we can see the certified ESDF region marked in green, and even as the robot moves around a significant part of the area around the robot remains part of the safe region. 

\section{Conclusions}

\subsection*{Limitations and Future Directions}

While the proposed methods are provably correct, they rely on key assumptions, particularly~\Cref{assumption:p}, which truncates the normal distribution of pose perturbations to bound the effects of a rototranslation on an obstacle point. Although this simplification facilitates our framework, it may not hold in practice. Methods such as those in~\cite{barfoot2024state, mangelson2020characterizing} could improve these approximations and warrant further exploration.

Additionally, we assumed that incremental odometry perturbations follow a normal distribution in the Lie algebra of $\SE(3)$. However, this may not hold in practice, especially with outliers (see e.g.~\cite{yang2020teaser}). A valuable direction for future work is to rigorously characterize the error distribution of odometry systems, both analytically and empirically.

We also highlight the need for modern perception algorithms to report the uncertainty of incremental pose transforms \edit{(e.g., as in~\cite{maken2021stein})}, rather than overall pose error/covariance, which grow unbounded without successful loop closures. Metrics such as relative translation and rotation errors~\cite{zhang2018tutorial} or the correlation between pose uncertainties (as in~\cite{mangelson2020characterizing}) should be computed and reported. In lieu of this, our experiments estimated incremental pose error covariances using the method described in~\Cref{appendix:extracting_relative_covariance}. For certifiability guarantees, going forward we will need odometry algorithms capable of directly reporting the incremental pose error covariance.

Our algorithm intentionally deflates the map, and this reduces the navigable volume for the robot. It is challenging to estimate how much the volume reduces prior to a mission, since the deflation depends on the exact obstacle geometry, features used by the odometry algorithm, and the speed of the robot (which affects how quickly new parts of the environment are observed). Empirically, we have shown that as the odometry covariance decreases, the volume of the free space increases, and approaches the volume of baseline methods in the error-free case~(\Cref{appendix:effect_of_covariance}). We also operated our rover in a larger room, and in~\Cref{appendix:collab_space} we show empirically that the certified methods can yield similar or larger volumes of free space than the heuristic method. Further analysis into this warranted. 

\edit{While this paper focused on deflating the map to ensure correctness, future work can consider methods to reinflate deflated regions when the correctness can be guaranteed again. For example, when a loop closure is detected, the odometry drift is reduced, and therefore uncertified regions can perhaps be marked as certifiably free again. To achieve this however we will require further analysis into the correctness guarantees of loop closures (e.g.~\cite{rosen2019se}), as well as efficient algorithms and map representations to handle the inflation and deflation steps.} 

Beyond odometry drift, there are other sources of error that can invalidate the correctness of the map - the operating environment and each subsystem can introduce errors that are hard to correct or even detect.  For instance, depth estimation algorithms (e.g., block-matching methods) can fail under conditions like glass surfaces or featureless walls. Similarly, communication/computational latencies can introduce errors that are hard to characterize with the current framework.

\subsection*{Summary}

As robots increasingly operate in unstructured environments, the importance of tightly integrated perception, planning, and control systems becomes evident. Our experiments demonstrate that even over short distances, perception inaccuracies due to odometry drift can lead to unsafe behaviors, including collisions.

This paper presents a step toward building perception modules that not only generate accurate state estimates and obstacle maps but also provide correctness guarantees. Specifically, if the incremental odometry error per frame can be bounded, our framework modifies (or deflates) obstacle-free regions in a map such that it remains correct at all times with respect to the robot’s body frame.

We proposed two methods for implementing these corrections based on different map representations: (I)~Certified~\acp{SFC}, and (II)~Certified~\acp{ESDF}. By constructively proving the correctness of these methods, we developed algorithms that guarantee safe map modifications. Extensive simulations using high-quality datasets, along with real-world experiments on a robotic rover, validate the effectiveness of our approach in creating certifiably-correct maps.

A key insight from our rover experiments is the demonstration of failure modes in state-of-the-art mapping methods. Unlike typical demonstrations, where robots map regions within the camera’s field of view or use 360$^\circ$ sensors (e.g., LIDAR), we intentionally operated the robot in its blind spot to highlight the challenges posed by accumulated odometry drift. Our proposed methods successfully mitigated these issues, preventing collisions and ensuring safe navigation.




\section*{Acknowledgments}
The authors would like to acknowledge the support of the National
Science Foundation (NSF) under grant no. 1942907. 


\bibliographystyle{ieeetr}
\bibliography{biblio}

\onecolumn

\appendix

\subsection{Review of Matrix Lie Groups}
\label{appendix:lie_groups}

Here we review the fundamentals of representing a pose and its uncertainty through the language of Lie groups and Lie algebras. We refer to readers to~\cite{sola2018micro, mangelson2020characterizing, barfoot2024state} and references therein for a more complete description.

The Lie group $\SO(3)$ is the set of valid 3D rotation matrices, and the group $\SE(3)$ is the set of rigid transformations in 3D:
\eqnN{
\SO(3) &= \left \{ R \in \R^{3 \times 3} : R R^T = I_3, \det{R} = 1 \right \}, \\
\SE(3) &= \left \{ T = \bmat{R & t\\ 0 & 1} \in \R^{4 \times 4} : R \in \SO(3), t \in \R^3 \right \}.
}

Both $\SO(3)$ and $\SE(3)$ are matrix Lie groups, i.e., the group composition operation is the standard matrix multiplication operation. 

The group action for $\SE(3)$ is $\cdot : \SE(3) \times \R^3 \to \R^3$, which transforms a point $p$ from its representation in frame $A$ to that in frame $B$. Given $\T{A}{B} = \bmat{R & t \\ 0 & 1} \in \SE(3)$, 
\eqn{
p\inframe{B} = \T{A}{B} \cdot p\inframe{A} = R p \inframe{A} + t.
}

The tangent space centered at identity is called the Lie algebra of a Lie group. The Lie algebra is a vector space of all possible directions an element of the group can be perturbed locally. The Lie algebras of $\SO(3)$ and $\SE(3)$ are denoted $\so(3)$ and $\se(3)$ respectively: 
\eqnN{
\so(3) &= \left \{ \omega \in \R^{3 \times 3} : \omega^T = - \omega  \right \},\\
\se(3) &= \left \{  \bmat{ \omega & \rho \\ 0 & 0} \in \R^{4 \times 4} : \omega \in \so(3), \rho \in \R^3 \right \}.
}

These vector spaces are isomorphic to the Euclidean vector space $\R^3$ and $\R^6$ respectively. The $\wedge$ operator converts the Euclidean vector to an element of the Lie Algebra. For $\SO(3)$, $\wedge: \R^3 \to \so(3)$:
\eqn{
\phi^\wedge = \bmat{\phi_1 \\ \phi_2 \\ \phi_3}^\wedge = \bmat{ 0 & -\phi_3 & \phi_2 \\ \phi_3 & 0 & -\phi_1 \\ -\phi_2 & \phi_1 & 0}
}
while for $\SE(3)$, $\wedge : \R^6 \to \se(3)$:
\eqn{
\xi^\wedge = \bmat{ \rho \\ \phi}^\wedge = \bmat{ \phi^\wedge & \rho \\ 0 & 0}.
}
The $\vee$ operator performs the inverse of $\wedge$.

Given an element of the Lie algebra, we can convert it to the corresponding element of the group using the exponential map. For $\SE(3)$, the exponential map is $\exp: \se(3) \to \SE(3)$, 
\eqn{
\exp(X) = \sum_{k=0}^\infty \frac{X^k}{k!} = I + X + \frac{X^2}{2} + \cdots
}
For convenience, we also define the $\Exp$ map, which maps from the Euclidean representation directly to the group element, $\Exp: \R^6 \to \SE(3)$, 
\eqn{
\Exp(\xi) = \exp(\xi^\wedge).
}
Analytic expressions for this are provided in~\cite[Appendix]{sola2018micro}. The corresponding inverse operations are $\log $ and $\Log$.

The adjoint matrix of $\SE(3)$ at $T \in \SE(3)$ is the unique matrix $\Ad{T} \in \R^{6 \times 6}$ such that  
\eqn{
T \Exp(\xi) = \Exp(\Ad{T} \xi) T
}
for all $\xi \in \R^6$. Again, the analytic expression is available in~\cite[Appendix]{sola2018micro}.

\newpage
\subsection{Proof of \Cref{lemma:polytope_separating}}
\label{appendix:proofs:polytope_separating}

Before we prove~\Cref{lemma:polytope_separating}, we derive a separating hyperplane result, \Cref{lemma:separating}. It defines the hyperplane that separates potential obstacle points from the free space after an uncertain rigid transformation.
\begin{lemma}
        \label{lemma:separating}
        Let the transform between two frames be $\T{A}{B} \sim \Ncal( \estT{A}{B}, \Sigma)$. 
        Consider a point $p\inframe{A} \in \R^3$. Given~\Cref{assumption:p}, for any non-zero vector $a \in \R^3$, 
        \eqn{
                p \inframe{B} &= \T{A}{B} \cdot p\inframe{A} \in \Hcal
                }
                where
                \seqn{
                \Hcal &= \{ p \in \R^3 : a^T p \geq r \}\\
                r &= a^T (\estT{A}{B} \cdot p\inframe{A} ) - \sqrt{a^T \Sigma_p a} \label{eqn:separating_r}
        }
        and $\Sigma_p \in \pd^{3}$ is as defined by~\Cref{assumption:p}. 
\end{lemma}
\begin{IEEEproof}[Proof of~\Cref{lemma:separating}]
        By~\Cref{assumption:p}, the transformed point satisfies
        \neqn{
                p\inframe{B} \in \Ecal =  \left\{ p \in \R^3 : \norm{ \Sigma_p^{-1/2} (p - \hat p)} \leq 1 \right\}
        }
        where $\hat p = \estT{A}{B} \cdot p\inframe{A}$, and $\Sigma_p \in \pd^3$ is defined in~\Cref{assumption:p}.
        Next, we define
        \neqn{
                p^{\perp} = \hat p - \frac{\Sigma_p a}{\sqrt{a^T \Sigma_p a}}
        }
        such that $p^{\perp} \in \R^3$ is on the surface of the ellipsoid and has a surface normal $-a$. 
        Therefore, the set of points $\Hcal = \{ p \in \R^3 : a^T(p - p^{\perp}) \geq 0\}$ contains the ellipsoid, i.e., $\Ecal \subset \Hcal$, 
        \neqn{
                r = a^T p^\perp = a^T \hat p - \frac{a^T \Sigma_p a}{\sqrt{a^T \Sigma_p a}}
                                = a^T \hat p - \sqrt{a^T \Sigma_p a}
        }
        which completes the proof. 
\end{IEEEproof}

We can now prove~\Cref{lemma:polytope_separating}.

\begin{IEEEproof}[Proof of~\Cref{lemma:polytope_separating}]
        It suffices to show that any obstacle potentially on the boundary of $\Pcal_k$ will not be in $\Pcal_{k+1}$. 
        Consider an obstacle point $o\inframe{B_{k}} = p\inframe{B_k} + \epsilon a_k$, where $\epsilon > 0$ and $p\inframe{B_k}$ is a point on the surface of $\Pcal_k$. Then for some $i \in \{1, ..., N\}$,
        \neqn{
                a_{k, i}^T p\inframe{B_k} = b_{k, i}.
        }
        
        After the rigid transformation, by~\Cref{lemma:separating}, $o\inframe{B_{k+1}} \in \Ecal \subset \{ p : a_{k+1, i}^T p \geq r\}$
        where 
        \neqn{
               r &= a_{k+1, i}^T ( \estT{B_k}{B_{k+1}} \cdot o\inframe{B_k}) - \sqrt{a_{k+1, i}^T \Sigma_p a_{k+1, i}}\\
                  &= a_{k+1, i}^T (R (p\inframe{B_k} + \epsilon a_{k, i}) + t) - \sqrt{a_{k+1, i}^T \Sigma_p a_{k+1, i}}\\
                  &= a_{k, i}^T (p\inframe{B_k} + \epsilon a_{k, i}) + a_{k, i}^T R^T t - \sqrt{a_{k+1, i}^T \Sigma_p a_{k+1, i}}\\
                  &= b_{k, i} + \epsilon \norm{ a_{k, i}}^2 + a_{k, i}^T R^T t - \sqrt{a_{k+1, i}^T \Sigma_p a_{k+1, i}}\\
                  &= b_{k+1, i} + \epsilon \norm{ a_{k, i}}^2 + \rho_i - \sqrt{a_{k+1, i}^T \Sigma_p a_{k+1, i}}
        }
        
        Now consider the last term:
        \neqn{
        \sqrt{a_{k, i+1}^T \Sigma_p a_{k+1, i}} &= \norm{ \Sigma_p^{1/2} a_{k+1, i}}\\
                &= \norm{  \sqrt{\kappa} \Sigma_k^{1/2} J^T a_{k+1, i}}\\
                &= \norm{  \sqrt{\kappa} \Sigma_k^{1/2} \bmat{ R^T \\ -(R [o\inframe{B_{k}}]_\times)^T} a_{k+1, i}}\\
                &= \norm{  \sqrt{\kappa} \Sigma_k^{1/2} \bmat{ a_{k, i} \\ [o \inframe{B_{k}}]_\times a_{k, i}}}\\
                &= \norm{  \sqrt{\kappa} \Sigma_k^{1/2} \bmat{ a_{k, i} \\ -[a_{k, i}]_\times o \inframe{B_{k}}}}\\
                &= \norm{  \sqrt{\kappa} \Sigma_k^{1/2} \bmat{ a_{k, i} \\ -[a_{k, i}]_\times p \inframe{B_{k}}}}
        }
        where in the last line, we used $[a_{k, i}]_\times (\epsilon a_{k, i}) = 0$. 
        
        Finally, since $\Sigma_k$ is positive definite, this expression is convex wrt $p\inframe{B_k}$. 
        Considering $p\inframe{B_k}$ must be some convex combination of the vertices on the $i$-th face, 
        \neqn{
                \norm{ \Sigma_p^{1/2} a_{k+1, i}} &\leq \max_{j \in \{1, ..., m_i\}} 
                \norm{  \sqrt{\kappa} \Sigma_k^{1/2} \bmat{ a_{k, i} \\ -[a_{k, i}]_\times v_{i, j}\inframe{B_k} }} \\
                &= \rho_i
        }
        where $v_{i,j}\inframe{B_k}$ is the $j$-th vertex on the $i$-th face of $\Pcal_k$. 
        
        Therefore, we have 
        \neqn{
                r &= b_{k+1, i} + \epsilon \norm{ a_{k, i}}^2 + \rho_i - \norm{ \Sigma_p^{1/2} a_{k+1, i}}\\
                  &\geq b_{k+1, i} + \epsilon \norm{a_{k, i}}^2 > b_{k+1, i},
        }
        that is, 
        \neqn{
                &o\inframe{B_{k+1}} \in \Ecal \subset \{ p : a_{k+1, i}^T p \geq r \}, \\
                &\implies o\inframe{B_{k+1}} \not \in \{ p : a_{k+1, i}^T p \leq b_{k+1, i}\}
        }
        which completes the proof. 
\end{IEEEproof}

\newpage
\subsection{Proof of \Cref{theorem:esdf_theorem}}
\label{appendix:proof:esdf_theorem}
\begin{IEEEproof}
        Consider any point $p\inframe{B_{k+1}}$. When represented in frame $B_k$, it could correspond to a set of points within the ellipsoid
        \neqn{
                p\inframe{B_k} \in \Ecal = \{ p \in \R^3: \norm{ \Sigma_p^{-1/2} ( p - \hat p)} \leq 1 \}
        }
        where $\hat p = \estT{B_{k+1}}{B_k} \cdot p\inframe{B_{k+1}}$, and $\Sigma_p \in \pd^3$ is as defined by~\Cref{assumption:p}.
        Therefore, 
        \neqn{
                d(p\inframe{B_{k+1}}) &\stackrel{(1)}{\geq} \min_{p\inframe{B_k} \in \Ecal} d(p\inframe{B_k})\\
                                      &\stackrel{(2)}{\geq} \min_{p\inframe{B_k} \in \Ecal} d_M^k( \estT{B_k}{M} \cdot p\inframe{B_k})\\
                                      &\stackrel{(3)}{\geq} d_M^k(\estT{B_k}{M} \cdot \hat p) - \operatorname{diam}(\Ecal)/2\\
                                      &\stackrel{(4)}{=} d_M^k(\estT{B_k}{M} \estT{B_{k+1}}{B_k} \cdot p\inframe{B_{k+1}}) - \sqrt{\lambda_{\max}(\Sigma_p)}\\
                                      &\stackrel{(5)}{=} d_M^k(\estT{B_{k+1}}{M} \cdot p\inframe{B_{k+1}}) - \sqrt{\lambda_{\max}(\Sigma_p)}\\
                                      &\stackrel{(6)}{=} d_M^{k+1} ( \estT{B_{k+1}}{M} \cdot p\inframe{B_{k+1}})
        }
        where $\operatorname{diam}(\Ecal)$ is the diameter of $\Ecal$. (1) is true by defition, (2) uses the fact that $d_M^k$ is a certified-\ac{ESDF}. (3) is true because \acp{ESDF} have unit gradient everywhere, (4) uses the eigenvalue of $\Sigma_p$ to bound the ellipsoid with a sphere, and (5), and (6) are basic simplifications. 
        Therefore, $d_M^{k+1}$ is also a certified-\ac{ESDF}.
\end{IEEEproof}

\newpage
\subsection{Extracting Covariance of Relative Transforms from Odometry with Covariance}
\label{appendix:extracting_relative_covariance}

To the best of the author's knowledge, all \ac{VO}/\ac{VIO}/\ac{SLAM} algorithms report the mean odometry estimate and the covariance with respect to the initial frame: at the $k$-th frame, the following quantities are available:
\eqn{
\estT{B_k}{B_0} \in \SE(3), \quad \Sigma_{B_k}^{B_0} \in \pd^6
}
i.e., the pose of the $k$-th body frame with respect to the initial frame, and the covariance of the estimate. 

However, to use the frameworks proposed in this paper,  the relative transform and its covariance are required:
\eqn{
\estT{B_{k+1}}{B_k} \in \SE(3), \quad \Sigma_{B_{k+1}}^{B_k} \in \pd^6.
}
Here we detail a method to obtain these quantities. 

Consider the following result adapted from~\cite[Section VIII]{mangelson2020characterizing} to match the convention used in this paper. 
\begin{lemma}
  \label{lemma:relative_covariance}  
    Let $T_{ij}, T_{ik}, T_{jk} \in \SE(3)$ represent the poses between coordinate frames $(i, j), (i, k)$, and $(j, k)$ respectively. Let $\hat T_{\cdot}$ be the corresponding estimated transform. Let
    \eqn{
    T_{ij} = \hat T_{ij} \Exp(\xi_{ij})
    }
    and similar for $(ik), (jk)$. Suppose 
    \eqn{
    \bmat{\xi_{ij}\\ \xi_{ik}} \sim \Ncal\left( \bmat{0 \\ 0}, \bmat{ \Sigma_{ij} & \Sigma_{ij, jk} \\ \Sigma_{ij, ik}^T & \Sigma_{ik}} \right).
    }
    Then, the estimated relative transform is 
    \eqn{
    \hat T_{jk} = \hat T_{ij}^{-1} \hat T_{ik}
    }
    and the associated covariance is (to first order)
    \eqn{
    \Sigma_{jk} = A \Sigma_{ij} A^T + \Sigma_{ik} - A \Sigma_{ij,ik} - \Sigma_{ij, jk}^T A^T,
    }
    where $A = \Ad{\hat T_{jk}^{-1}} \in \R^{6 \times 6}$ is the adjoint matrix of $\SE(3)$ at $\hat T_{jk}^{-1}$. 
\end{lemma} 
 Notice that the negative signs on the cross terms implies that  a non-zero $\Sigma_{ij, jk}$ decreases the covariance of the relative pose. 
\begin{IEEEproof}
Since  $T_{jk} = T_{ij}^{-1} T_{ik}$, the following must hold:
\eqnN{
\hat T_{jk} \Exp(\xi_{jk}) &= \left( \hat T_{ij} \Exp(\xi_{ij}) \right)^{-1} \left( \hat T_{ik} \Exp(\xi_{ik})\right)\\
&= \Exp(-\xi_{ij}) \hat T_{ij}^{-1} \hat T_{ik} \Exp(\xi_{ik})\\
&= \Exp(-\xi_{ij}) \hat T_{jk} \Exp(\xi_{ik})\\
&= \hat T_{jk} \Exp( - \Ad{\hat T_{jk}^{-1}} \xi_{ij}) \Exp( \xi_{ik})
}
where in the last equality we used  the following property of the adjoint matrix: $\Exp(\xi) T = T \Exp( \Ad{T^{-1}} \xi)$ for any $T \in \SE(3)$ and $\xi \in \R^6$. 

Defining $\xi_{ij}' = -\Ad{\hat T_{jk}^{-1}} \xi_{ij}$, we have
\eqnN{
\Exp(\xi_{jk}) = \Exp(\xi_{ij}')\Exp(\xi_{ik})
}
and therefore using the \ac{BCH} formula (see~\cite{mangelson2020characterizing}), the first order estimated covariance is 
\eqnN{
E[\xi_{jk}\xi_{jk}^T] 
&\approx \underbrace{E[\xi_{ij}' \xi_{ij}'{}^T]
+ E[\xi_{ik}\xi_{ik}^T]}_{\text{2nd order diag. terms}}\\
&\quad
+ \underbrace{E[\xi_{ij}' \xi_{ik}^T]
+ E[\xi_{ik}\xi_{ik}'{}^T]}_{\text{2nd order cross terms}}\\
&= A \Sigma_{ij} A^T + \Sigma_{ik} - A \Sigma_{ij,ik} - \Sigma_{ij, jk}^T A^T
}
where $A = \Ad{\hat T_{jk}^{-1}}$. This completes the proof. 
\end{IEEEproof}

We can now apply this lemma to estimate the relative transforms between successive frames. 
Recall the odometry algorithm defines the covariances as
\eqn{
\T{B_k}{B_0} = \estT{B_k}{B_0} \Exp(\xi_{k, 0}), \quad \xi_{k, 0} \sim \Ncal(0, \Sigma_{k, 0})
} 
and similar for $k+1$. The perturbations $\xi$ are assumed to be correlated,
\eqn{
\bmat{\xi_{k, 0} \\ \xi_{k+1, 0}} \sim \Ncal \left(  \bmat{0 \\ 0}, 
\bmat{ \Sigma_{k, 0}   &  \Sigma_{k, 0; k+1, 0} \\ * & \Sigma_{k+1, 0}} \right)
}
where the $*$ indicates to the symmetric element. 

We assume that the two poses are highly correlated, with a correlation coefficient $\rho \in [-1, 1]$, (we chose $\rho = 0.99$). Then, 
\eqn{
\Sigma_{k, 0; k+1, 0} = \rho \left( \Sigma_{k, 0} \Sigma_{k+1, 0}^T \right)^{1/2}
}

Then, using \Cref{lemma:relative_covariance},  the estimated relative transform is 
\eqn{
\estT{B_{k+1}}{B_{k}} = (\estT{B_k}{B_0})^{-1} \estT{B_{k+1}}{B_{0}}
}
and the estimated relative covariance is 
\eqn{
\Sigma_{B_{k+1}}^{B_{k}} = A \Sigma_{B_k}^{B_0} A^T + \Sigma_{B_{k+1}}^{B_0} - A \Sigma_\times - \Sigma_\times^T A^T
}
where 
\eqnN{
\Sigma_\times = \rho \left( \Sigma_{B_k}^{B_{0}} (\Sigma_{B_{k+1}}^{B_{0}})^T \right)^{1/2}, \quad
A = \Ad{(\estT{B_{k+1}}{B_{k}})^{-1}}.
}

Note, the adjoint matrix for $T = \bmat{R & t \\ 0 & 1} \in \SE(3)$ is  
\eqnN{
\Ad{T} = \bmat{ R & [t]_\times R \\ 0 & R}
}
and $\Ad{T^{-1}} = (\Ad{T})^{-1}$~\cite{sola2018micro}.

\newpage
\subsection{Replica Dataset Environment Details}

\Cref{tab:bounding_boxes} shows the size and volume of the bounding box for each environment used in the simulation studies. It also shows the number of mesh points in the environment. 

\begin{table*}[h!]
\centering
\caption{Size and volume of each environment used.}
\label{tab:bounding_boxes}
{%
\begin{tabular}{@{}l S[table-format=1.2] S[table-format=1.2]S[table-format=1.2]S[table-format=3.2]S[table-format=8.0]@{}}
\toprule
Env. & {Length X (m)} & {Length Y (m)} & {Length Z (m)} & {Bounding Box Volume (m$^3$)} & {Number of Mesh Points}\\
\midrule
\texttt{office0} & 4.40 & 5.01 & 2.99 & 65.95   & 589517 \\
\texttt{office1} & 4.81 & 4.11 & 2.80 & 55.24   & 423007 \\
\texttt{office2} & 6.47 & 8.14 & 2.77 & 145.89  & 858623 \\
\texttt{office3} & 8.64 & 9.20 & 3.10 & 246.85  & 1187140 \\
\texttt{office4} & 6.55 & 6.51 & 2.82 & 119.96  & 993008 \\
\texttt{room0}   & 7.76 & 4.70 & 2.81 & 102.43  & 954492 \\
\texttt{room1}   & 6.65 & 5.73 & 2.75 & 104.81  & 645512 \\
\texttt{room2}   & 6.77 & 4.95 & 3.59 & 120.34  & 722496 \\
\bottomrule
\end{tabular}%
}
\end{table*}

\newpage
\subsection{Additional Simulation Results}
\label{appendix:additional}

\Cref{table:sfc_results_additional} and \Cref{table:sdf_results_additional} show additional results of the performance of the \ac{SFC} and \ac{ESDF} methods on the Replica dataset. Here we show the results from a trajectory perturbed by $\Sigma=$1e-5$I$ and $\Sigma=$1e-6$I$. 

\begin{table*}[h!]
\centering
\caption{Results of the three \acf{SFC} methods on the Replica dataset. Each environment was run with $\Sigma = \sigma^2 I$ for two different $\sigma^2$ values, 1e-5 and 1e-6.}
\label{table:sfc_results_additional}
{%
\begin{tabular}{@{}cr|rrg|rrg|rrg@{}}
        \toprule
                    &     & \multicolumn{3}{c|}{Violation Rate (\%)} & \multicolumn{3}{c|}{Max Violation (mm)} & \multicolumn{3}{c}{SFC Volume (m$^3$)} \\
   Env & $\sigma^2$&
  Baseline & Heuristic & Certified& 
  Baseline & Heuristic & Certified& 
  Baseline & Heuristic & Certified\\
  \midrule
\texttt{office0}&  1e-6    & 18.6\pct    & 0.1\pct    & 0.0\pct   & 102.74    & 22.05    & 0.03   & 34.8   & 6.7   & 5.7   \\
                &  1e-5    & 32.5\pct    & 0.6\pct    & 0.0\pct   & 397.89    & 33.83    & 0.03   & 38.9   & 6.8   & 4.8   \\
\texttt{office1}&  1e-6    & 12.8\pct    & 0.6\pct    & 0.0\pct   & 95.30     & 14.48    & 0.86   & 17.6   & 3.6   & 2.6   \\
                &  1e-5    & 12.9\pct    & 0.1\pct    & 0.0\pct   & 373.39    & 24.65    & 0.86   & 17.7   & 3.7   & 2.0   \\
\texttt{office2}&  1e-6    & 10.1\pct    & 0.1\pct    & 0.0\pct   & 159.66    & 18.42    & 0.39   & 40.8   & 4.3   & 3.6   \\
                &  1e-5    & 21.3\pct    & 0.9\pct    & 0.0\pct   & 299.11    & 21.93    & 0.39   & 44.9   & 4.3   & 3.0   \\
\texttt{office3}&  1e-6    & 12.7\pct    & 0.1\pct    & 0.0\pct   & 177.65    & 11.61    & 0.88   & 56.6   & 4.6   & 3.0   \\
                &  1e-5    & 16.5\pct    & 0.0\pct    & 0.0\pct   & 460.25    & 7.38     & 0.94   & 57.9   & 4.6   & 0.9   \\
\texttt{office4}&  1e-6    & 14.4\pct    & 0.3\pct    & 0.0\pct   & 125.48    & 8.91     & 1.69   & 63.3   & 15.7  & 12.5  \\
                &  1e-5    & 24.6\pct    & 4.7\pct    & 0.0\pct   & 262.23    & 82.75    & 1.69   & 66.5   & 16.1  & 10.6  \\
\texttt{room0}  &  1e-6    & 10.7\pct    & 0.0\pct    & 0.0\pct   & 117.12    & 11.02    & 0.95   & 53.0   & 12.3  & 9.1   \\
                &  1e-5    & 20.1\pct    & 0.5\pct    & 0.0\pct   & 396.74    & 47.97    & 0.95   & 55.8   & 12.3  & 8.0   \\
\texttt{room1}  &  1e-6    & 19.2\pct    & 0.4\pct    & 0.0\pct   & 191.43    & 14.20    & 0.71   & 38.7   & 6.9   & 5.8   \\
                &  1e-5    & 25.7\pct    & 1.1\pct    & 0.0\pct   & 377.01    & 23.68    & 0.71   & 39.5   & 6.7   & 5.3   \\
\texttt{room2}  &  1e-6    &  6.8\pct    & 0.9\pct    & 0.0\pct   &  85.02    & 12.85    & 0.65   & 29.4   & 7.5   & 4.4   \\
                &  1e-5    & 11.1\pct    & 1.5\pct    & 0.0\pct   & 322.36    & 25.63    & 0.65   & 30.1   & 7.5   & 1.8  \\
                         \bottomrule
\end{tabular}%
}
\end{table*}

\begin{table*}[h!]
\centering
\caption{Results of the three \acf{ESDF} methods on the Replica dataset.}
\label{table:sdf_results_additional}
{%
\begin{tabular}{@{}cr|rrg|rrg|rrg@{}}
        \toprule
                    &     & \multicolumn{3}{c|}{Violation Rate (\%)} & \multicolumn{3}{c|}{Max Violation (mm)} & \multicolumn{3}{c}{ESDF Volume (m$^3$)} \\
   Env & $\sigma^2$&
  Baseline & Heuristic & Certified& 
  Baseline & Heuristic & Certified& 
  Baseline & Heuristic & Certified\\
  \midrule
\texttt{office0}&  1e-6   & 48.1\pct & 31.6\pct & 0.5\pct & 604.3 & 563.6 & 109.5 & 46.1  & 39.5 & 10.7 \\
                &  1e-5   & 21.2\pct & 11.8\pct & 0.5\pct & 384.2 & 322.5 & 107.7 & 42.3  & 38.1 & 10.9 \\
\texttt{office1}&  1e-6   & 35.3\pct & 34.4\pct & 0.1\pct & 406.9 & 379.5 & 82.5  & 23.2  & 23.0 & 3.8  \\
                &  1e-5   & 11.1\pct & 10.6\pct & 0.3\pct & 172.0 & 172.0 & 93.8  & 21.9  & 21.8 & 4.2  \\
\texttt{office2}&  1e-6   & 51.5\pct & 7.6\pct  & 0.1\pct & 520.0 & 311.8 & 141.4 & 77.5  & 31.3 & 6.2  \\
                &  1e-5   & 23.8\pct & 2.0\pct  & 0.1\pct & 212.6 & 253.8 & 100.0 & 68.7  & 31.1 & 6.2  \\
\texttt{office3}&  1e-6   & 54.7\pct & 4.7\pct  & 0.0\pct & 671.1 & 429.4 & 100.0 & 110.9 & 42.0 & 5.0  \\
                &  1e-5   & 28.2\pct & 1.5\pct  & 0.0\pct & 330.5 & 226.3 & 72.1  & 96.9  & 41.4 & 6.0  \\
\texttt{office4}&  1e-6   & 48.3\pct & 10.1\pct & 0.1\pct & 636.9 & 366.6 & 66.3  & 99.7  & 51.5 & 14.3 \\
                &  1e-5   & 21.0\pct & 3.9\pct  & 0.1\pct & 260.0 & 215.4 & 69.3  & 90.9  & 50.9 & 14.4 \\
\texttt{room0}  &  1e-6   & 62.0\pct & 9.2\pct  & 2.4\pct & 990.8 & 428.5 & 120.0 & 105.4 & 28.6 & 31.5 \\

                &  1e-5   & 34.4\pct & 3.2\pct  & 3.2\pct & 335.3 & 244.1 & 164.9 & 90.9  & 27.6 & 32.9 \\
\texttt{room1}  &  1e-6   & 48.1\pct & 20.9\pct & 0.0\pct & 604.6 & 384.7 & 100.0 & 53.8  & 34.5 & 6.6  \\
                &  1e-5   & 17.5\pct & 8.8\pct  & 0.0\pct & 240.0 & 169.7 & 72.1  & 47.6  & 33.1 & 6.9  \\
\texttt{room2}  &  1e-6   & 47.5\pct & 16.3\pct & 0.1\pct & 594.0 & 435.4 & 82.5  & 63.6  & 38.7 & 4.5  \\
                &  1e-5   & 21.9\pct & 5.1\pct  & 0.0\pct & 291.9 & 200.0 & 66.3  & 56.8  & 37.8 & 9.5 \\
                         \bottomrule
\end{tabular}%
}
\end{table*}

\newpage
\subsection{Effect of Odometry Covariance}
\label{appendix:effect_of_covariance}

\begin{table*}[h!]
\centering
\caption{Performance of the three \acf{ESDF} methods in the Office0 environment under varying odometry covariance at $\kappa=3$.}
\label{table:effect_of_odometry_covariance}
{%
\begin{tabular}{@{}r|rrg|rrg|rrg@{}}
        \toprule
                         & \multicolumn{3}{c|}{Violation Rate (\%)} & \multicolumn{3}{c|}{Max Violation (mm)} & \multicolumn{3}{c}{ESDF Volume (m$^3$)} \\
    $\sigma^2$&
  Baseline & Heuristic & Certified& 
  Baseline & Heuristic & Certified& 
  Baseline & Heuristic & Certified\\
  \midrule

1e-04 & 61.09\pct  & 43.46\pct & 0.19\pct  &  1.24  &  1.24  &  0.17  &  63.87  &  45.37  &  10.68\\
1e-05 & 48.15\pct  & 31.55\pct & 0.50\pct  &  0.60  &  0.56  &  0.11  &  46.14  &  39.46  &  10.74\\ 
1e-06 & 21.16\pct  & 11.79\pct & 0.49\pct  &  0.38  &  0.32  &  0.09  &  42.33  &  38.11  &  10.94\\
1e-07 & 5.17\pct   & 2.75\pct  & 0.49\pct  &  0.22  &  0.19  &  0.10  &  41.79  &  37.93  &  11.59\\
1e-08 & 2.04\pct   & 1.72\pct  & 0.54\pct  &  0.18  &  0.13  &  0.13  &  41.70  &  37.86  &  16.54\\
1e-09 & 1.94\pct   & 1.54\pct  & 0.62\pct  &  0.18  &  0.12  &  0.16  &  41.69  &  37.85  &  31.52\\
1e-10 & 1.91\pct   & 1.51\pct  & 0.77\pct  &  0.18  &  0.11  &  0.16  &  41.69  &  37.85  &  37.18\\
1e-11 & 1.93\pct   & 1.53\pct  & 1.21\pct  &  0.18  &  0.11  &  0.16  &  41.69  &  37.86  &  41.20\\
1e-12 & 1.93\pct   & 1.53\pct  & 1.35\pct  &  0.18  &  0.11  &  0.16  &  41.69  &  37.86  &  41.59\\

\bottomrule
\end{tabular}%
}
\end{table*}

\begin{figure*}[h!]
    \centering
        \includegraphics[width=1\linewidth]{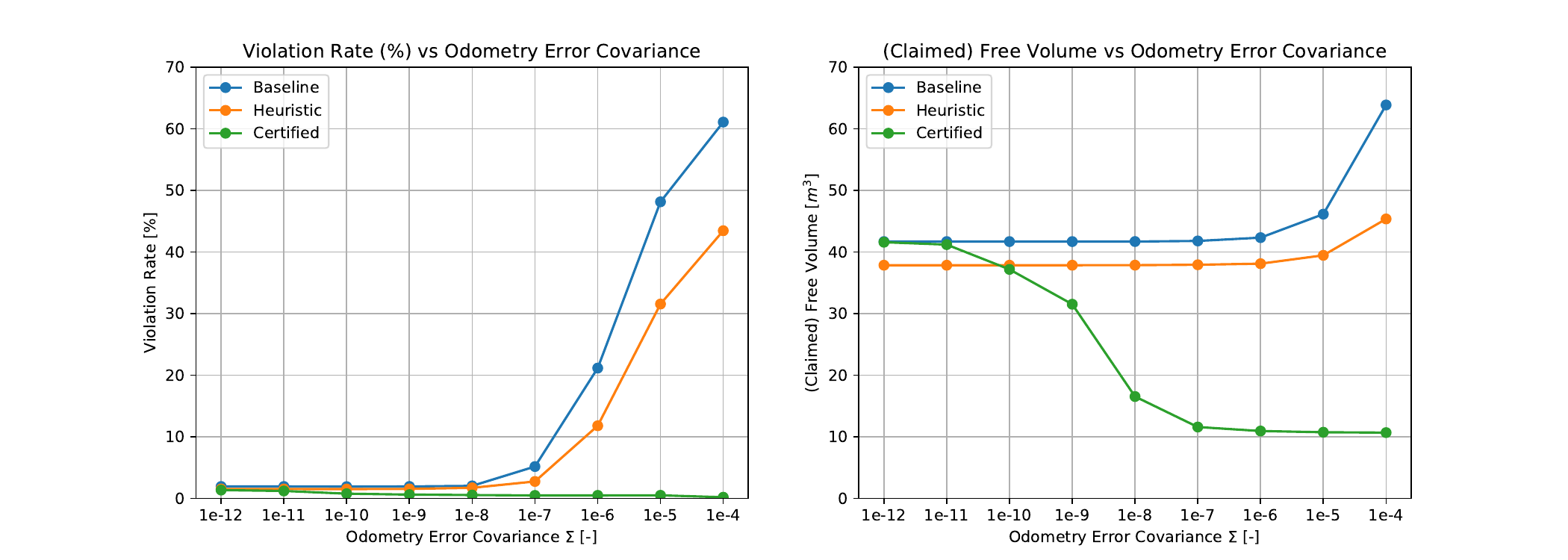}
    \caption{(Left) Effect of the odometry covariance on the mapping violation rate. (Right) Effect of the odometry covariance on the claimed free volume. Notice that the true free volume is approximately 42~m$^3$, and in the uncertified methods the volume of claimed free space incorrectly increases beyond 42~m$^3$. In contrast, in the certified methods the volume decreases to reflect the increased uncertainty.    }
    \label{fig:odometry_covariance_effect}
\end{figure*}

\end{document}